\documentclass{article}
\usepackage[utf8]{inputenc}
\usepackage{amsmath}
\usepackage{amsfonts}
\usepackage{graphicx}
\usepackage{booktabs}
\usepackage{amssymb}
\usepackage{ragged2e}
\usepackage{geometry}
\usepackage{float}
\usepackage{longtable} 
\usepackage{verbatim} 
\usepackage{tabularx}
\usepackage[table]{xcolor}
\newcolumntype{C}{>{\centering\arraybackslash}X}
\usepackage{subcaption}

\usepackage{bm}
\usepackage{tikz}
\usetikzlibrary{
  arrows.meta,
  positioning,
  shapes.geometric,
  calc,
  fit,
  backgrounds,
  quotes, 
  decorations.pathreplacing 
}
\tikzset{
  every node/.style={inner sep=2pt}
}

\definecolor{inputGreen}{RGB}{225, 245, 238}
\definecolor{inputGreenBorder}{RGB}{15, 110, 86}
\definecolor{encoderGray}{RGB}{241, 239, 232}
\definecolor{encoderGrayBorder}{RGB}{95, 94, 90}
\definecolor{computeBlue}{RGB}{230, 241, 251}
\definecolor{computeBlueBorder}{RGB}{24, 95, 165}
\definecolor{stagePurple}{RGB}{238, 237, 254}
\definecolor{stagePurpleBorder}{RGB}{83, 74, 183}
\definecolor{projOrange}{RGB}{250, 236, 231}
\definecolor{projOrangeBorder}{RGB}{153, 60, 29}
\definecolor{codebookYellow}{RGB}{250, 238, 218}
\definecolor{codebookYellowBorder}{RGB}{133, 79, 11}
\definecolor{deadGreen}{RGB}{234, 243, 222}
\definecolor{deadGreenBorder}{RGB}{59, 109, 17}

\title{Probing the Latent World: Emergent Discrete Symbols and Physical Structure in Latent Representations}

\author{Liu Hung Ming\thanks{PARRAWA AI} \\ \texttt{cyril.liu@gmail.com}}

\begin{document}

\maketitle

\begin{abstract}
Video world models trained with Joint Embedding Predictive
Architectures (JEPA) acquire rich spatiotemporal representations by
predicting masked regions in latent space rather than reconstructing
pixels.
This design produces powerful encoders, but removes the visual
verification pathway available to generative models, creating a
structural interpretability gap: the encoder has learned physical
structure, but that structure is not accessible in any inspectable
form.
Existing probing methods either operate in continuous space without
introducing a structured intermediate layer, or attach learned
generative components whose own parameters confound the attribution
of observed behavior to the encoder's representations.

We propose attaching the \textbf{AI Mother Tongue (AIM)} framework as
a \emph{passive quantization probe}: a lightweight probe with minimal
task-specific inductive bias that converts V-JEPA~2's continuous latent
vectors into discrete symbol sequences without task-specific supervision
or predefined symbol inventory, and without modifying the encoder.
Because the encoder is kept completely frozen throughout, any symbolic
structure that emerges in the AIM codebook is attributable entirely to
V-JEPA~2's pre-trained representations---not to the probe.

We evaluate this design through category-contrast experiments on Kinetics-mini, contrasting action-category pairs along
three physical dimensions: grasp angle, object geometry, and motion
temporal structure.
We use action-category pairs as proxies for physical dimensions, following a category-contrast strategy; direct physical manipulation of individual variables is deferred to Stage~3.
AIM symbol distributions differ significantly across all three
category-contrast experiments ($\chi^2\ p < 10^{-4}$;
absolute mutual information $0.036$--$0.117$\,bits,
normalized MI $1.2$--$3.9\%$ of the $\log_2 8 = 3$\,bit theoretical maximum;
Jensen--Shannon divergence up to $0.342$) and codebook utilization is healthy
($62.5\%$ active entries).

Beyond confirming architectural compatibility between AIM and
V-JEPA~2, the experiments reveal a characteristic property of
V-JEPA~2's latent space: diverse action categories share a common
representational core, with semantic differences encoded as graded
distributional variations rather than categorical boundaries.
This compactness is strongest along temporal structure dimensions and
weakest along static morphological dimensions, consistent with
V-JEPA~2's temporal prediction objective.
We argue this reflects the model's success in internalizing shared
physical structure---not a limitation of representational
capacity---and that AIM's discrete symbol distributions provide a
statistically testable interface for characterizing it.

These results establish Stage~1 of a four-stage integration roadmap
toward an action-conditioned symbolic world model, and demonstrate
that structured symbolic manifolds are discoverable properties of
frozen JEPA latent spaces rather than artefacts of training-time
discretization objectives.
\end{abstract}

\section{Introduction}
\label{sec:intro}

Contemporary self-supervised video models have achieved substantial
progress by operating entirely in latent space.
The Joint Embedding Predictive Architecture
(JEPA)~\cite{lecun2022path}, instantiated for video in
V-JEPA~\cite{bardes2024vjepa} and extended in
V-JEPA~2~\cite{assran2025vjepa2}, trains a visual encoder to predict
the latent representation of masked spatiotemporal regions rather than
reconstructing pixels.
This objective encourages the encoder to internalize the physical
regularities of the visual world---object kinematics, scene geometry,
temporal continuity---while remaining agnostic to surface-level
appearance details that are irrelevant to prediction.
The result is a representational system of considerable empirical
power: V-JEPA~2 achieves state-of-the-art performance on motion
understanding (77.3\% top-1 on Something-Something-v2) and human
action anticipation (39.7 recall-at-5 on Epic-Kitchens-100), and
enables zero-shot robot manipulation planning without task-specific
training or reward~\cite{assran2025vjepa2}.

Yet the same design that makes JEPA-style models powerful introduces a structural interpretability challenge. Unlike generative models that reconstruct pixels and thereby provide a visual verification pathway, JEPA deliberately confines all learning and prediction to latent space~\cite{lecun2022path}. This is not incidental opacity---it is a necessary consequence of the architecture: the model has no built-in mechanism for reading its own representations back out into an inspectable form.

A researcher can verify that V-JEPA~2 performs well on a downstream
task, but cannot identify which physical concepts are encoded, how
they are organized, or whether the representations support the kind of
structured reasoning that the world-model hypothesis
implies~\cite{lecun2022path}.
We refer to this as the \textbf{representational opacity} problem: the
encoder has acquired structured knowledge, but that structure is not
accessible in a form that permits scientific audit.

\paragraph{Limitations of existing interpretability approaches.}
Current methods for probing neural representations fall into two broad
categories, each with a fundamental limitation.

\emph{Discriminative probes}---linear classifiers, nonlinear decoders,
and representation similarity analyses~\cite{alain2017probing,
tenney2019bert}---establish that a particular variable can be decoded
from the latent space.
However, they do not introduce an intermediate representational layer;
they produce a yes/no answer about decodability rather than a
structured interface that can be independently analyzed or communicated
across systems.
Moreover, they operate in a continuous space and therefore cannot
provide the kind of discrete, auditable symbolic record that
interpretability ultimately requires.

\emph{Generative probes}---language model heads, pixel decoders, and
task-specific generative components attached to the encoder
output---translate latent representations into human-readable form.
But these approaches introduce a confound that is rarely acknowledged:
when the composite system performs well, it is impossible to determine
how much of the observed behavior originates in the encoder's
representations and how much is contributed by the attached
component's own learned parameters.
A language model head attached to a frozen encoder can produce fluent,
accurate descriptions by drawing on its own linguistic priors, even
when the encoder's representations carry little structured information
about the relevant concepts.
The encoder and the generative head are not separable in the
inferential sense.
We term this the \textbf{attribution problem}: interpretability
methods that rely on learned generative components cannot cleanly
attribute their outputs to the model under study.

\paragraph{This work: passive discrete probing.}
We propose a different approach.
Rather than attaching a learned generative component, we introduce the
\textbf{AI Mother Tongue (AIM)} framework~\cite{liu2025aim} as a
\emph{passive quantization probe}: a lightweight probe with minimal
task-specific inductive bias that converts V-JEPA~2's continuous latent
vectors into sequences of discrete symbols without task-specific
supervision or predefined symbol inventory, and without modifying the
encoder in any way.
The AIM quantizer carries the inductive biases inherent to vector
quantization (cluster-forcing geometry, codebook size, EMA update
dynamics), but imposes no semantic labels, language supervision, or
task-specific objectives that could produce symbolic structure
independently of the encoder.

AIM was originally developed as a mechanism through which agents in
multi-agent reinforcement learning develop a self-emergent compressed
semantic language~\cite{liu2025aim}.
The AIM framework~\cite{liu2025aim} operates by attaching a VQ-VAE
bottleneck~\cite{vandenoord2017vqvae} to an agent's latent
communication channel, converting continuous latent vectors into
discrete symbol sequences through nearest-neighbor assignment to a
learned codebook.
The resulting symbols are not predefined but emerge from the statistics
of the agent's internal representations.
In the present work, we apply this mechanism as a read-only probe
attached to a frozen encoder rather than as an active communication
channel---its core operation is therefore architecturally compatible
with V-JEPA~2 because both frameworks operate entirely in latent space.
V-JEPA~2 predicts in latent space; AIM compresses in latent space.
No pixel decoder, no language model, and no task-specific head is
required or used.

The key inferential property of this design is that the V-JEPA~2
encoder is kept \emph{completely frozen} throughout: every parameter
gradient is blocked and the encoder runs in evaluation mode.
Because the mapping from input to latent vector is therefore fixed and
deterministic, any symbolic structure that emerges in the AIM codebook
is attributable entirely to V-JEPA~2's pre-trained representations.
The encoder cannot adapt to help the quantizer, and the quantizer
cannot import semantic structure that the encoder does not already
possess.
This resolves the attribution problem: a positive finding is evidence
about V-JEPA~2, not about the probe.
The formal justification and full architectural specification of this
design are given in Sections~\ref{sec:framework}
and~\ref{sec:methods}.

\paragraph{The central question.}
We ask a precise and falsifiable question:

\begin{quote}
\emph{Does the frozen latent space of V-JEPA~2 already contain
structured manifolds that a vocabulary-free discrete probe can detect
and statistically characterize under controlled physical-dimension contrast?}
\end{quote}

We investigate this through \emph{category-contrast experiments}: we select action-category pairs that contrast sharply
along one physical dimension (grasp angle, object geometry, motion
temporal structure) while minimizing differences along other salient
visual factors, and we measure whether the AIM symbol distribution
shifts in a statistically significant manner.
Statistical evidence is provided by mutual information (MI),
the Jensen--Shannon divergence (JSD), and the chi-squared test, with a
Gaussian-noise baseline anchoring the null distribution.
These metrics allow us to evaluate symbolic structure without relying
on human annotation or downstream task accuracy~\cite{hewitt2019structural,
alain2017probing}.

\paragraph{Scope.}
In this work, we focus on a minimal and controlled instantiation:
we apply AIM as a probing layer on a frozen V-JEPA~2 encoder and
evaluate whether the induced symbolic distributions correlate with
controlled physical variables.
We make no claim that the discrete symbols constitute a complete
semantic representation of the physical world, nor that they establish
causal understanding in the encoder.
The goal of this stage is diagnostic: to verify architectural
compatibility between AIM and V-JEPA~2, and to establish whether the
frozen latent space contains extractable structure sufficient to
motivate subsequent joint-training stages.

\paragraph{Contributions.}
This paper makes the following contributions.

\begin{enumerate}

  \item \textbf{Passive discrete probing.}
    We introduce and formalize the distinction between passive probing
    (frozen encoder, vocabulary-free discrete probe) and active probing
    (learned generative attachments), and argue that passive probing
    provides a cleaner causal basis for attributing symbolic structure
    to the model under study rather than to the probe.

  \item \textbf{Architectural compatibility.}
    We demonstrate that AIM~\cite{liu2025aim} can be attached to a
    frozen V-JEPA~2~\cite{assran2025vjepa2} encoder without
    modification of any original source files, and that a lightweight
    single-layer VQ quantizer~\cite{vandenoord2017vqvae} trains stably
    on pre-computed spatial token vectors extracted from the frozen
    encoder.

  \item \textbf{Statistically significant symbolic structure.}
    Controlled category-contrast experiments across three physical
    dimensions yield symbol distributions that differ significantly
    between conditions across all three interventions
   ($\chi^2\ p < 10^{-4}$; absolute MI $0.036$--$0.117$\,bits,
    normalized MI $1.2$--$3.9\%$ of the $3$\,bit theoretical maximum;
    JSD up to $0.342$), establishing that
    V-JEPA~2's frozen latent space encodes physically structured
    information that is recoverable through discrete symbolization.

  \item \textbf{Compact latent-space characterization.}
    The experiments reveal that V-JEPA~2's latent space is markedly
    compact: diverse action categories share a common representational
    core, with semantic differences encoded as graded distributional
    variations rather than categorical boundaries.
    We argue this reflects the model's success in internalizing the
    shared physical structure underlying surface-level
    variation~\cite{lecun2022path,assran2025vjepa2}---a property
    consistent with world-model training objectives---rather than a
    failure of representational capacity.

\end{enumerate}

\paragraph{Paper organization.}
Section~\ref{sec:framework} introduces the conceptual framework
separating the latent model, discrete semantic, and language interface
layers, and defines the scope of the present work within that
framework.
Section~\ref{sec:related} reviews related work in latent
representation learning, interpretability and probing, and discrete
representation learning~\cite{baek2025discrete,wu2024nlotm}, positioning
the present work relative to prior approaches.
Section~\ref{sec:methods} describes the dataset, frozen encoder
design, quantizer architecture, and category-contrast experiment protocol.

Sections~\ref{sec:experiments} presents the
Stage~1 experimental results and analysis.
Section~\ref{sec:discussion} discusses the passive-probe methodology,
the compact latent-space finding, and the limitations of the
category-proxy strategy.
Section~\ref{sec:future} outlines the four-stage integration roadmap
and positions Stage~1 within the broader AIM--JEPA research program.

\section{Conceptual Framework}
\label{sec:framework}

\subsection{Overview}

Many contemporary models---including predictive architectures, world
models, and reinforcement learning agents---operate by encoding
observations into a latent space where internal representations are
formed, updated, and used for prediction or control.
These latent representations are effective precisely because they
abstract away from surface-level sensory detail and capture the
underlying structure of the environment.
But this abstraction comes at a cost: the representations are
continuous, high-dimensional, and carry no built-in semantic
vocabulary.
The gap between internal model states and human-understandable
meaning is not a secondary concern; it is a structural feature of the
architecture.

We propose a conceptual framework that addresses this gap by
introducing a discrete intermediate layer between a latent model and
any downstream interpretability or communication interface.
The framework separates three distinct functions---representation,
discretization, and interpretation---into three corresponding layers,
each with a well-defined role and a well-defined interface to its
neighbours.
This separation is not merely organizational.
It has a specific inferential consequence that is central to the
present work: by keeping the layers decoupled, any symbolic structure
that emerges at the discrete layer can be attributed to the latent
model rather than to the interpretation layer.

\subsection{The Three-Layer Architecture}
\label{sec:three_layers}

We formalize the system as a composed transformation over three
layers.
The three layers and their interfaces are described below.

\paragraph{Layer 1: Latent Model Layer.}
A model $E_\phi$ maps input observations $\mathbf{x}$ into a
continuous latent representation:
\begin{equation}
  \mathbf{z} = E_\phi(\mathbf{x}) \in \mathbb{R}^D.
  \label{eq:encoder}
\end{equation}
In the present work, $E_\phi$ is the V-JEPA~2
encoder~\cite{assran2025vjepa2}, a Vision Transformer pretrained to
predict masked spatiotemporal regions in latent space.
The encoder is kept frozen throughout: $\nabla_\phi \mathcal{L} = 0$.
This layer is not restricted to video encoders; the framework applies
to any model that produces a continuous latent representation,
including reinforcement learning world models and predictive coding
architectures.

\paragraph{Layer 2: Discrete Semantic Layer (AIM).}
A vector-quantization module $Q_\psi$ maps the continuous
representation $\mathbf{z}$ to a discrete symbol from a finite
codebook $\mathcal{C} = \{e_1, \ldots, e_K\}$:
\begin{equation}
  s = Q_\psi(\mathbf{z})
    = \operatorname*{arg\,min}_{k} \| \mathbf{z} - e_k \|_2,
  \qquad s \in \{1, \ldots, K\}.
  \label{eq:quantizer}
\end{equation}
The codebook entries $\{e_k\}$ are learned from data; the symbol
assignments $s$ are determined entirely by the geometry of the latent
space.
Because $Q_\psi$ introduces no task-specific supervision or predefined
symbol inventory---it does not know what categories exist, what
language labels apply, or what physical variables are present---any
systematic relationship between the resulting symbols and the physical
world must originate in $E_\phi(\mathbf{x})$, not in $Q_\psi$ itself.
This property is the foundation of the inferential logic described in
Section~\ref{sec:intro}.

The discrete semantic layer is implemented using the AIM
framework~\cite{liu2025aim}, originally developed for emergent
communication in multi-agent reinforcement learning.
AIM's vocabulary-free design makes it directly applicable here: the same mechanism that allows MARL agents to develop a shared symbolic language from latent representations can be used to probe whether a
pre-trained latent model has already encoded structured semantic content. Vocabulary-free denotes the absence of predefined category labels or language supervision; the vector quantization mechanism itself carries inductive biases including cluster-forcing geometry and codebook size.

\paragraph{Layer 3: Language Interface Layer.}
A language model $\mathcal{L}$ maps sequences of discrete symbols
into natural language, enabling human interpretation of the internal
states recorded by Layer~2:
\begin{equation}
  \text{description} = \mathcal{L}(s_1, s_2, \ldots, s_T).
  \label{eq:language}
\end{equation}
This layer is the decompression interface: it translates the
compressed symbolic record produced by Layer~2 into a form that
humans can read and reason about.
Unlike generative probes that are attached directly to the latent
model output, a language model operating on discrete symbol sequences
receives no access to the continuous latent representation $\mathbf{z}$.
Any interpretation it produces is therefore grounded in the symbolic
record rather than in the raw latent vectors, which constrains the
degree to which the language model can substitute its own priors for
genuine representational content.

\subsection{The Role of Discretization}
\label{sec:role_discretization}

The central architectural choice in this framework is the placement of
a discrete bottleneck between the latent model and the interpretation
interface.
This choice has consequences that go beyond data compression.

First, discretization converts a continuous, unstructured latent space
into a finite symbolic vocabulary.
Individual symbols are reusable and comparable across inputs: if video
clip $A$ and video clip $B$ both map to symbol $s = 5$, this
constitutes a claim---falsifiable by statistical test---that the two
clips are similar along whatever dimension the codebook has learned to
track.
Continuous latent representations do not support this kind of
auditable claim without additional machinery.

This property---that vector quantization renders latent content into auditable statistical objects---has been independently exploited in the context of multi-agent safety, where it enables detection of covert coordination in agent communication channels~\cite{liu2026drcb} (in preparation).

Second, a discrete bottleneck makes the information flow inspectable.
The symbol sequence $s_1, \ldots, s_T$ is a complete record of what
passed through the bottleneck, and its statistical properties
(distribution, mutual information with external variables, temporal
patterns) can be measured directly.
This is the basis for the category-contrast experiments in
Section~\ref{sec:methods}: we measure whether symbol distributions
shift when physical conditions change, using standard statistical
tests that require no human annotation.

Third, discretization decouples the representational layer from the
interpretation layer in the inferential sense described above.
A continuous latent vector passed directly to a language model can be
shaped by the language model's own parameters; a symbol sequence
passed to a language model carries only the information that survived
the bottleneck.

\subsection{Scope of the Present Work}
\label{sec:scope}

The three-layer framework is intended as a general interface
applicable to a wide class of latent-variable models, including video
encoders, reinforcement learning agents, and multi-agent world models.
However, the present work instantiates only a minimal configuration:

\begin{itemize}
  \item \textbf{Layer 1} is a frozen V-JEPA~2 ViT-L
    encoder~\cite{assran2025vjepa2}; the encoder weights are never
    modified.
  \item \textbf{Layer 2} is a lightweight single-layer VQ
    quantizer~\cite{vandenoord2017vqvae} trained only on
    pre-computed latent vectors from the frozen encoder; there is no
    joint training between layers.
  \item \textbf{Layer 3} is not instantiated in this work.
    The present study stops at the discrete symbol level and does not
    attach a language model.
\end{itemize}

The goal of this minimal instantiation is diagnostic rather than
generative.
We do not attempt to build a complete interpretable system.
We ask only whether the frozen latent space contains structure that is
recoverable through discretization---a prerequisite that must be
established before the full three-layer pipeline can be meaningfully
assembled.

\subsection{Positioning: What This Framework Does and Does Not Claim}
\label{sec:positioning}

We treat V-JEPA~2 as a latent representation system that has learned
structure sufficient for spatiotemporal prediction, rather than as a
complete or fully grounded model of the physical world.
Under this view, the discrete semantic layer does not recover
``true semantics'' in any philosophical sense.
What it produces is a \emph{structured interface}: a symbolic record
of regularities in the latent space that correlate with physical
conditions and that can be statistically audited.

Two claims that this framework explicitly does \emph{not} make are
worth stating clearly.

First, we do not claim that V-JEPA~2 is a world model in the sense
of LeCun~\cite{lecun2022path}---a system capable of causal reasoning,
planning, and counterfactual prediction.
The evidence we present is consistent with that hypothesis but does
not establish it.
What we can say is that V-JEPA~2's representations exhibit properties
that a world model \emph{should} have: they encode physically grounded variables
in a compact, shared form that is sensitive to temporal dynamics.

Second, we do not claim that the symbolic distinctions produced by
Layer~2 are semantically interpretable without further work.
A symbol assigned the index $s = 5$ has no intrinsic meaning; its
meaning, if any, must be established by examining which physical
conditions consistently map to it.
The statistical tests in Section~\ref{sec:methods} establish that
such mappings exist and are significant; they do not establish what
the symbols \emph{mean}.
That question is deferred to later stages of the research program
described in Section~\ref{sec:future}.

\section{Related Work}
\label{sec:related}

Having established the three-layer framework in
Section~\ref{sec:framework}, we now situate the present work within
the existing literature.
We organize prior work into three lines of research, each
corresponding to one of the three layers in our framework, and
conclude by articulating the specific gap that the present work
addresses.

\subsection{Latent Representation Learning and JEPA}

The Joint Embedding Predictive Architecture
(JEPA)~\cite{lecun2022path} proposes that intelligent systems should
model the world by predicting representations in a learned abstract
space rather than reconstructing raw sensory inputs.
This framing explicitly prioritizes semantic structure over perceptual
fidelity, and motivates an encoder design that captures physical
regularities---object motion, spatial layout, causal
continuity---without being distracted by stochastic, uninformative
surface details.

V-JEPA~\cite{bardes2024vjepa} instantiates this principle for video,
training a Vision Transformer encoder to predict masked spatiotemporal
regions in latent space using a feature prediction objective that
requires no pretrained image encoders, text supervision, or
contrastive negatives.
V-JEPA~2~\cite{assran2025vjepa2} scales this approach to over one
million hours of internet video, achieving state-of-the-art
performance on motion understanding and human action anticipation, and
demonstrating zero-shot robot manipulation planning through a
post-trained action-conditioned predictor.

A central property of JEPA-style models that motivates the present
work is their deliberate absence of a pixel decoder.
Unlike masked autoencoders~\cite{he2022mae} or video generation
models that produce pixel-level reconstructions, JEPA encodes and
predicts entirely in latent space.
This design produces more semantically structured representations, but
simultaneously removes the visual verification pathway that generative
models provide: there is no reconstructed output through which an
observer can inspect what the encoder has learned.
The interpretability challenge addressed in this paper is therefore
not incidental to JEPA's design---it is a structural consequence of
it.

\subsection{Interpretability and Latent Probing}

A substantial body of work has developed methods for interpreting
neural representations through probing.
Linear classifier probes~\cite{alain2017probing} assess whether a
target variable can be decoded from a frozen representation using a
simple linear model; the probe's accuracy is taken as evidence that
the variable is linearly encoded.
More structured probing approaches~\cite{tenney2019bert} apply this
logic to specific linguistic or semantic properties, revealing which
layers of a deep network encode which types of information.

These methods have provided valuable insights into the internal
organization of language models and, to a lesser extent, vision
models.
However, they share two limitations that are relevant to the present
work.

First, discriminative probes operate in continuous space and produce
a scalar accuracy score rather than a structured intermediate
representation.
They answer the question ``is variable $X$ decodable from this
layer?'' but do not construct a reusable symbolic interface that can
be independently analyzed, compared across inputs, or communicated
between systems.

Second, and more fundamentally, attaching a learned probe introduces
what we term the attribution problem: when a probe performs well, it
is unclear whether the performance reflects genuine structure in the
probed representation or the probe's own capacity to fit the target
variable.
Hewitt and Liang~\cite{hewitt2019structural} address a related
concern by introducing control tasks to measure probe complexity, but
the core confound---that a sufficiently expressive probe can succeed
even on unstructured representations---remains.
Our passive quantization approach resolves this confound by design:
the AIM quantizer imposes no task-specific supervision or predefined
symbol inventory, and the frozen encoder cannot adapt to the probe.

\subsection{Discrete Representation Learning and Neural Symbolic
Abstraction}
Vector Quantized Variational Autoencoders
(VQ-VAE)~\cite{vandenoord2017vqvae} introduced the idea of learning
discrete codebook representations through end-to-end training,
demonstrating that neural networks can develop reusable symbolic units
without explicit symbolic supervision.
This work established the technical foundation for the discrete
semantic layer in our framework.

Wu et al.~\cite{wu2024nlotm} proposed Neural Language of Thought
Models (NLoTM), demonstrating that a Semantic VQ-VAE can learn
discrete codes corresponding to independent semantic factors in
images.
Through factor-level quantization and latent traversal, their results
show that discrete symbols can capture disentangled and causally
meaningful structure---but this capability relies on end-to-end
training explicitly designed to produce such representations.

Baek et al.~\cite{baek2025discrete} introduced Discrete JEPA,
showing that vector quantization can be integrated into the JEPA
training objective to produce discrete tokens that support symbolic
reasoning.
Their results provide strong evidence that JEPA latent spaces are
compatible with symbolic abstraction when a discretization objective
is incorporated into training.

The AIM framework~\cite{liu2025aim}, developed in the context of
multi-agent reinforcement learning, takes a different approach:
rather than training agents to produce discrete symbols through
task-specific objectives, AIM demonstrates that a VQ-VAE bottleneck
allows agents to develop a self-emergent symbolic language from their
latent representations without external inductive biases.
This vocabulary-free property is what makes AIM directly applicable
as a passive probe: it does not require a predefined symbol inventory
or a semantic supervision signal.

A further application of AIM's discretization principle appears in
the DRCB framework~\cite{liu2026drcb} (in preparation), where vector quantization is
used to convert unobservable latent communication between agents into
auditable statistical objects, enabling detection of covert
coordination.
That work independently demonstrates the core property exploited in
the present paper: that discretization renders latent content
inspectable and statistically testable, regardless of the domain in
which the latent model operates.

\subsection{Positioning of This Work}

The three lines of research above converge on a common gap.
JEPA-style models produce powerful latent representations but provide
no interpretability interface.
Discriminative probes can detect whether variables are encoded but
cannot construct a structured intermediate layer, and they introduce
the attribution problem.
Discrete representation methods demonstrate that symbolic structure
can be learned or induced, but focus on training-time objectives
rather than on probing pre-existing structure in frozen models.

This work addresses that gap by asking a more constrained question:
does a frozen, pre-trained latent model already contain structured
symbolic manifolds that a vocabulary-free discrete probe can reveal?
The design differs from prior work in three respects.

\begin{itemize}
  \item \textbf{No joint training.}
    The latent encoder is frozen throughout; the quantizer learns
    only from the fixed output distribution of the encoder.
    Any symbolic structure that emerges is therefore attributable to
    the encoder's pre-trained representations rather than to
    co-adaptation between the encoder and the probe.
    This contrasts with Discrete JEPA~\cite{baek2025discrete} and
    NLoTM~\cite{wu2024nlotm}, which demonstrate that symbolic
    representations \emph{can be learned}; our findings indicate
    that such structure \emph{can also be revealed} without modifying
    the underlying model.

 \item \textbf{Vocabulary-free discretization.}
    The AIM quantizer imposes no task-specific supervision or predefined
    symbol inventory on the latent space.\footnote{%
      \emph{Vocabulary-free} denotes the absence of predefined category labels
      or language supervision; the vector quantization mechanism itself carries
      inductive biases including cluster-forcing geometry and codebook size.}
    Codebook entries are initialized randomly and updated purely from
    data; no category labels, language annotations, or task-specific
    objectives are used.
    This resolves the attribution problem that affects both
    discriminative probes and generative interpretation methods.

\item \textbf{Causal evaluation via category-contrast experiments.}
    Rather than evaluating on downstream task accuracy, we directly
    compare action-category pairs along specified physical dimensions
    and measure whether symbol distributions shift in a statistically
    significant manner.
    This category-contrast evaluation strategy provides a more 
    direct test of whether the latent space encodes specific physical
    structure than accuracy on aggregate benchmarks.
\end{itemize}

Together, these design choices establish a methodologically cleaner
basis for attributing symbolic structure to the world model itself,
rather than to any component introduced during the probing procedure.

\section{Methods} \label{sec:methods}
\subsection{Dataset and Experimental Design}\label{sec:dataset}
\begin{figure}[H]
  \centering
\begin{tikzpicture}[
    font=\sffamily,
    node distance=1.2cm and 0.8cm,
    box/.style={rectangle, draw, rounded corners=6pt, inner sep=6pt, align=center, line width=0.6pt, minimum height=1.5cm},
    subbox/.style={rectangle, draw, rounded corners=4pt, inner sep=5pt, align=center, line width=0.5pt, minimum height=1.3cm, minimum width=2.8cm},
    arrow/.style={-{Stealth[scale=1.0]}, draw=gray!80, line width=1pt}
]

    \node[box, fill=encoderGray, draw=encoderGrayBorder, minimum width=3.5cm] (encoder) {
        \textbf{\small V-JEPA 2 ViT-L} \\ \scriptsize frozen $\nabla\theta = 0$ \\ \scriptsize $z \in \mathbb{R}^{B \times 1568 \times 1024}$
    };
    
    \node[box, fill=inputGreen, draw=inputGreenBorder, left=of encoder, minimum width=3.5cm] (input) {
        \textbf{\small Kinetics-mini} \\ \scriptsize 50 videos $\times$ 5 classes
    };

    \node[box, fill=computeBlue, draw=computeBlueBorder, right=of encoder, minimum width=3.5cm] (compute) {
        \textbf{\small Pre-compute z} \\ \scriptsize 75,264 token vectors
    };

    \draw[arrow] (input) -- (encoder);
    \draw[arrow] (encoder) -- (compute);

    \node[subbox, fill=codebookYellow, draw=codebookYellowBorder, below=2.5cm of encoder] (codebook) {
        \textbf{\small EMA Codebook} \\ \scriptsize K = 8 codewords \\ \scriptsize $\gamma = 0.90$ decay
    };

    \node[subbox, fill=projOrange, draw=projOrangeBorder, left=of codebook] (proj) {
        \textbf{\small Linear projection} \\ \scriptsize 1024 $\to$ 256 dim \\ \scriptsize + LayerNorm + L2
    };

    \node[subbox, fill=deadGreen, draw=deadGreenBorder, right=of codebook] (dead) {
        \textbf{\small Dead code reset} \\ \scriptsize every 200 steps \\ \scriptsize $\beta = 2.0$ commit loss
    };

    \draw[arrow] (proj) -- (codebook);
    \draw[arrow] (codebook) -- (dead);

    \node[fit=(proj) (dead), rectangle, draw=stagePurpleBorder, fill=stagePurple, fill opacity=0.3, rounded corners=12pt, inner sep=20pt ] (stageA) {};
    \node[anchor=north, text=stagePurpleBorder!80!black, yshift=-1pt] at (stageA.north) {\textbf{\small Stage A --- Quantizer training (3000 steps, encoder frozen)}};

    \node[left=0.8cm of stageA.west, text=gray!80] (labelA) {\textbf{\small A}};

    \draw[arrow] (input.south) -- (input.south |- stageA.north);
    \draw[arrow] (encoder.south) -- (encoder.south |- stageA.north);

    \node[subbox, fill=inputGreen, draw=inputGreenBorder, below=1.2cm of proj] (h1) {
        \textbf{\small H1: Symbol stability} \\ \scriptsize 20 repeats per video \\ \scriptsize consistency = 100\%
    };

    \node[subbox, fill=computeBlue, draw=computeBlueBorder, below=1.2cm of codebook] (h2) {
        \textbf{\small H2: Interventions} \\ \scriptsize 3 physical variables \\ \scriptsize $\chi^2$ + MI + JSD
    };

    \node[subbox, fill=encoderGray, draw=encoderGrayBorder, below=1.2cm of dead] (baseline) {
        \textbf{\small Random baseline} \\ \scriptsize Gaussian noise input \\ \scriptsize MI $\approx$ 0 (reference)
    };

    \node[at=(h1 -| labelA), text=gray!80] {\textbf{\small B/C}};

    \draw[arrow] (proj.south) -- (h1.north);
    \draw[arrow] (codebook.south) -- (h2.north);
    \draw[arrow] (dead.south) -- (baseline.north);

    \node[box, fill=deadGreen, draw=deadGreenBorder, below=1.5cm of h2, minimum width=10cm, minimum height=1.2cm] (output) {
        \textbf{\small Stage 1 report + AIM dictionary} \\ \scriptsize All Stage 1 criteria satisfied; Stage 2 entry conditions met };

    \draw[arrow] (h1.south) -- (output.165);
    \draw[arrow] (h2.south) -- (output.north);
    \draw[arrow] (baseline.south) -- (output.15);

\end{tikzpicture}
\caption{Stage~1 pipeline overview.
         \textit{Top row} (A): The input pipeline feeds 50 Kinetics-mini
         videos through the frozen V-JEPA~2 ViT-L encoder
         ($\nabla\theta = 0$), producing $B \times 1568 \times 1024$
         latent token vectors that are precomputed once and cached.
         \textit{Middle row} (A): The Stage~A quantizer trains for 3,000
         steps on the precomputed vectors using a linear projection layer
         ($1024 \to 256$ dimensions, followed by LayerNorm and L2
         normalization), an EMA-updated codebook ($K = 8$,
         $\gamma = 0.90$), and a dead-code reset mechanism
         ($\beta = 2.0$ commitment loss); the V-JEPA~2 encoder weights
         remain fully frozen throughout.
         \textit{Bottom row} (B/C): After training, the pipeline is
         evaluated on two diagnostic criteria: H1 symbol stability
         (20 repeated forward passes per video; consistency $= 100\%$)
         and H2 category-contrast experiments (three physical-variable pairs
         assessed via $\chi^2$, MI, and JSD), with a Gaussian-noise
         random baseline confirming MI $\approx 0$ for unstructured
         inputs.  All four pass criteria are satisfied, producing the
         Stage~1 report and AIM dictionary that certify readiness for
         Stage~2.}
  \label{fig:pipeline_overview}
\end{figure}
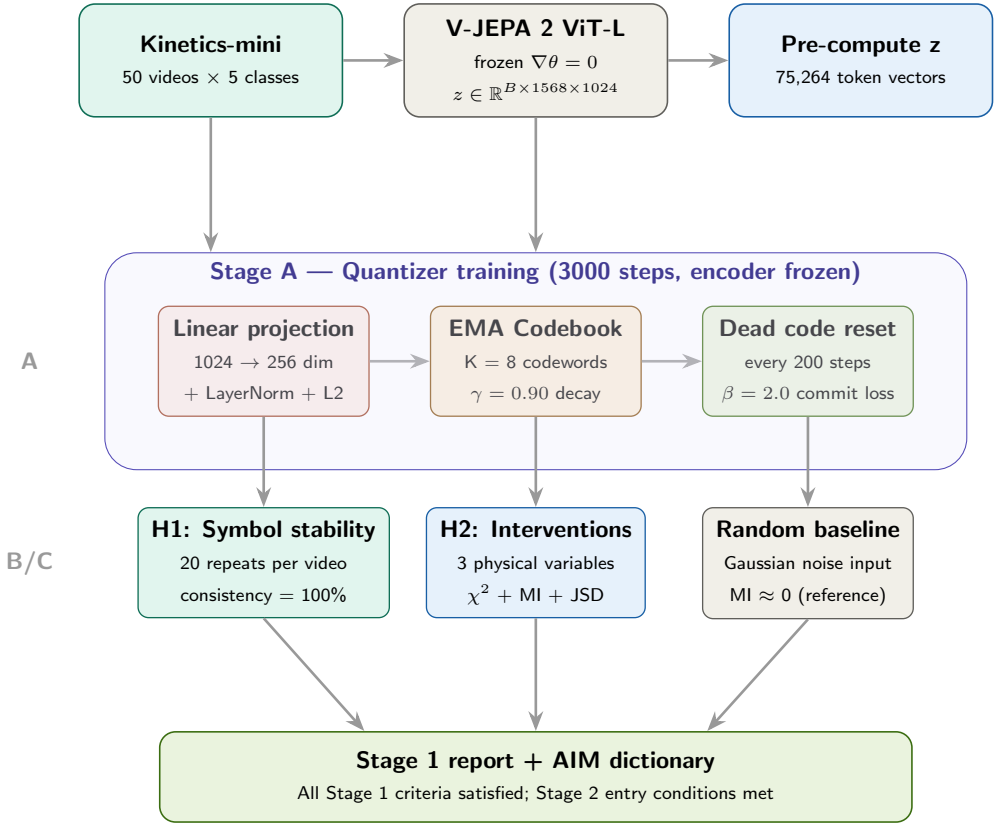

Figure~\ref{fig:pipeline_overview} provides an overview of the
complete Stage~1 pipeline, from data ingestion through quantizer
training to diagnostic evaluation.

\paragraph{Dataset.}
We use the validation split of the Kinetics-mini dataset, selecting five action
categories: \textit{archery}, \textit{bowling}, \textit{flying\_kite},
\textit{high\_jump}, and \textit{marching}, with at most 10 videos per category.
Of these, 48 videos ultimately entered training (see Section~\ref{sec:frozen_encoder} for details).
Each video is sampled at $T = 16$ frames at $224\times224$ resolution, processed
through V-JEPA~2's official processor with center cropping and ImageNet
normalization before being passed to the encoder.

\paragraph{Rationale for the contrast experiment design.}
The central question of Stage~1 is whether AIM's symbolization mechanism can
read structured information from V-JEPA~2's frozen latent space.  To investigate
this, we design controlled category-contrast experiments: all other conditions are
held fixed while a single physical dimension is varied, and we observe whether
the AIM symbol distribution changes in a statistically significant manner.

Kinetics-mini is an action-classification dataset whose labels denote action
categories rather than physical attributes.  It is therefore impossible to
directly manipulate a single physical variable (e.g., holding scene, subject,
and action constant while varying only grasp angle).  Instead, we adopt a
\textbf{category-proxy} strategy: we select pairs of action categories that
exhibit large natural differences along one target physical dimension while
minimizing differences along other major visual dimensions, using inter-category
contrast as a proxy for physical-variable contrast.

The validity of this strategy rests on the following premise.  V-JEPA~2 is
pretrained to predict masked spatiotemporal regions in latent space, so its
encoder must compress video into structured representations that support
spatiotemporal prediction.  Such representations inevitably encode physical
attributes such as object shape, pose, and motion pattern.  When two action
categories differ substantially along one physical dimension, that difference
should leave a measurable directional shift in latent space.  Stage~1 tests
whether this shift is sufficiently pronounced to be captured by AIM's discrete
symbolization mechanism.  Kinetics-mini serves as a publicly reproducible
benchmark providing a verifiable starting point.

Two practical reasons motivate the use of an existing video dataset rather than
purpose-built controlled footage.  First, the goal of this stage is
\emph{diagnostic}: to verify architectural compatibility between AIM and
V-JEPA~2, not to precisely quantify the encoding strength of specific physical
variables.  Category proxies are sufficient for the statistical-significance
tests that serve this purpose.  Second, acquiring video data with fine-grained
control over individual physical variables requires dedicated data-collection
infrastructure that exceeds the resource scope of Stage~1.

\paragraph{Category Pair Selection.}
The five categories are grouped into three contrastive pairs, each targeting one
physical dimension.  Pairs were chosen to maximize inter-category differences
along the target dimension while minimizing differences along other salient
visual features (e.g., scene type, number of persons, range of motion), thereby
reducing confounding variables.

\begin{itemize}
  \item \textbf{Grasp angle.}
    \textit{Archery} (drawing a bow: three-finger pinch grip on the string,
    arm extended into a static release posture with shoulder rotation)
    vs.\ \textit{bowling} (single-hand grip through finger holes, dynamic
    forward-swing throw).
    Both involve a single person, a hand-held object, and upper-limb-dominated
    motion; the primary differences are grip morphology and the direction of
    arm force application.  We note that scene environment (outdoor archery
    range vs.\ indoor bowling lane) and background color also differ---an
    unavoidable confound of the category-proxy strategy.

  \item \textbf{Object geometry.}
    \textit{Flying\_kite} (an elongated linear object controlled indirectly via
    string) vs.\ \textit{high\_jump} (no manipulated object; the subject's body
    executes an arching leap).
    The two conditions differ markedly in the geometry of the primary object and
    in the spatial relationship between the human body and that object.

  \item \textbf{Motion speed / temporal structure.}
    \textit{Marching} (periodic, regular gait with a $\approx$2\,Hz repetitive
    temporal structure) vs.\ \textit{archery} (near-static loading followed by a
    single rapid release; aperiodic temporal structure).
    Because V-JEPA~2's pretraining objective directly involves temporal
    prediction, we expect it to be most sensitive to temporal-structure
    differences---a prediction that is consistent with the experimental results,
    in which the motion-speed intervention yields the highest MI and JSD of the
    three pairs (Section~\ref{sec:h2_results}).
\end{itemize}

\paragraph{Reuse of Archery.}
\textit{Archery} appears in both the grasp-angle pair (contrasted with bowling)
and the motion-speed pair (contrasted with marching).  This is a design
compromise imposed by the limited dataset scale, but it simultaneously provides
an additional validation opportunity: if AIM's symbol system captures stable
latent features of the action category rather than incidental statistical
artifacts introduced by the intervention framework, then \textit{archery} should
produce consistent symbol distributions across the two independent interventions.
The experimental results confirm this (Section~\ref{sec:h2_results}).

\paragraph{Limitations of the Category-Proxy Strategy.}
The fundamental limitation of this approach is that the symbol-distribution
differences we observe reflect the overall differences between the selected
action categories in V-JEPA~2's latent space, not the isolated effect of the
target physical variable.  For the grasp-angle intervention, for instance, the
latent-space distance between \textit{archery} and \textit{bowling} subsumes
contributions from grasp angle, scene environment, object material, and other
factors that the experimental design cannot disentangle.  The correct
characterization of our results is therefore: \emph{AIM symbols have
statistically significant discriminative power for action-category pairs whose
primary difference lies along a specified physical dimension}---not that AIM
symbols directly encode the target physical variable.  Ideal follow-up
validation should employ carefully controlled synthetic or robot-manipulation
videos that isolate the target variable while holding all other visual factors
constant; this is deferred to Stage~3 or future work.

\subsection{Frozen V-JEPA~2 Encoder and Latent Space Precomputation}\label{sec:frozen_encoder}

\paragraph{Scientific Necessity of the Freezing Strategy.}
The core scientific question of this stage is whether AIM's symbolization
mechanism can read knowledge already acquired by V-JEPA~2.  If the encoder were
left unfrozen, two confounds would be introduced: (1) the encoder could
actively adapt its output distribution in response to the quantizer, and
(2) the quantizer could acquire separability through encoder gradients that
would not otherwise exist.  Freezing the encoder ensures that, for any input
$\mathbf{x}$, the mapping $E_\phi(\mathbf{x})$ is constant and all parameter
gradients are fully blocked:
\begin{equation}
  \forall\,\theta \in \Phi:\quad
  \frac{\partial \mathcal{L}}{\partial \theta} = 0,
  \qquad E_\phi(\cdot)\big|_{\text{eval}}.
\end{equation}
Freezing also yields a practical benefit: because a frozen encoder produces
identical outputs for identical inputs, all latent vectors $\mathbf{z}$ can be
precomputed once and reused throughout training, eliminating the need to
re-execute an encoder forward pass at every training step.

\paragraph{Preliminary Latent Space Diagnosis.}
Prior to designing the quantizer, we executed
 to characterize the numerical
properties of V-JEPA~2's output distribution across 5
representative videos.  This diagnostic revealed the key
challenge: the per-token L2 norm of $97.70 \pm 0.81$ is
exceptionally large and highly uniform (coefficient of
variation $< 0.6\%$), meaning that all latent vectors are
nearly equidistant from any randomly initialized codebook
entry.  Direct quantization without normalization would
therefore cause immediate codebook collapse, a finding that
directly motivated the projection-plus-normalization pipeline
described in Section~\ref{sec:quantizer}.
\paragraph{Token Structure of V-JEPA~2 ViT-L.}
V-JEPA~2 uses tubelet embedding with temporal patch size $t_p = 2$ and spatial
patch size $p = 16$.  For $T = 16$ input frames the number of output tokens is
\begin{equation}
  N_{\text{tokens}}
  = \frac{T}{t_p} \times \left(\frac{H}{p}\right)^2
  = \frac{16}{2} \times \left(\frac{224}{16}\right)^2
  = 8 \times 196
  = 1{,}568.
\end{equation}
Each tubelet token spans 2 frames in time and a $16\times16$ pixel patch in
space.  With embedding dimension $D = 1{,}024$, the encoder output for a batch
of $B$ videos is
\begin{equation}
  \mathbf{Z} = E_\phi(\mathcal{V}) \in \mathbb{R}^{B \times 1568 \times 1024}.
\end{equation}
Measurements on 5 representative videos show that the L2 norm of the latent
vectors satisfies $\|\mathbf{z}\|_2 = 97.70 \pm 0.81$ (range
$[96.26,\,98.58]$), a coefficient of variation below $0.6\%$.  This high
consistency is a systematic property of V-JEPA~2 rather than a coincidence;
however, the absolute magnitude is so large that direct quantization inevitably
distorts distance computations---all vectors are nearly equidistant from any
codebook entry.

\paragraph{Spatial Token Retention (Discarding Temporal Pooling).}
An earlier experimental version applied temporal pooling:
\begin{equation}
  \mathbf{z}_{\text{frame},t}
  = \frac{1}{196}\sum_{k=1}^{196}\mathbf{Z}_{t,k}
  \in \mathbb{R}^{1024},
\end{equation}
averaging the 196 spatial tokens for each time step.  This operation reduced
codebook utilization to only 5\% and caused the grasp-angle intervention to
fail statistical significance ($p = 0.357$).  The root cause is that mean
pooling erases spatial local semantic differences: the distributional contrast
between hand-pose tokens and background tokens is diluted into a single
per-frame average.

The final design retains all spatial tokens:
\begin{equation}
  \mathbf{z}_{\text{all}}
  = \operatorname{reshape}\!\bigl(\mathbf{Z},\;
    [B \cdot 1568,\; 1024]\bigr).
\end{equation}
The validity of this design presupposes that V-JEPA~2's latent vectors are
semantically diverse across the spatial dimension.  The 1,568 tokens are not
homogeneous: each corresponds to a distinct $16\times16$ pixel patch and carries
independent local semantics (hand pose, background texture, target-object shape,
etc.).  Temporal pooling actively destroys this diversity; retaining spatial
tokens allows the quantizer to learn patch-level semantic clusters directly.

In practice, the DataLoader scanned 50 videos with \texttt{drop\_last=True} and
batch size 16, so precomputation used
$\lfloor 50/16\rfloor \times 16 = 48$ videos, yielding
$48 \times 1{,}568 = 75{,}264$ token vectors---a 94-fold increase over the
800 vectors produced by temporal pooling.

Retaining raw spatial tokens allows different codebook entries to correspond to
local concepts such as ``hand motion,'' ``static background,'' and ``dynamic
foreground,'' rather than uninterpretable global averages.  This explains the
improvement in codebook utilization from 5\% (temporal pooling) to 62.5\%
(spatial tokens): the former provides training samples that are too homogeneous,
whereas the latter supplies the semantic diversity that codebook learning
requires.

\subsection{Stage~1 Lightweight AIM Quantizer (Stage~A)}\label{sec:quantizer}

\paragraph{Architecture Overview.}
The quantizer is a single-layer vector quantization (VQ) module; the only
trainable components are the projection layer $\mathbf{W}_p$ and the
EMA-updated codebook $\mathcal{C}$.  The single-layer design is deliberate:
the goal of Stage~1 is diagnostic---to verify architectural compatibility
between AIM and V-JEPA~2---and a single-layer VQ already provides sufficient
signal for the H2 statistical tests.  Residual quantization is deferred to
Stage~2.
\paragraph{Implementation Note.}
The Stage~1 quantizer described in this section is implemented as the
\texttt{Stage1AIMQuantizer} class.
A separate multi-level residual VQ architecture (\texttt{AIMQuantizerForVJEPA}
, $L = 3$ levels with codebook sizes
$\{64, 128, 256\}$) is reserved for Stage~2 and is not used in any
experiment reported here.  All hyperparameters reported in this section
(Table~\ref{tab:ablation}: $\gamma = 0.90$, $\beta = 2.0$, $K = 8$)
were passed as command-line arguments to,
overriding the class-level defaults retained for backward compatibility.
Stage 1 results are derived solely from the primary diagnostic framework. The Stage 2-ready architecture lacks the essential normalization pipeline; as such, it is incompatible with the frozen latent space under the current diagnostic constraints and would lead to a total loss of codebook utilization if applied here.

\paragraph{Projection Health Verification.}
Before committing to the full training run, we verified the
projection layer's behavior,
which confirmed that: (1) LayerNorm successfully stabilizes the
projected norm to $\sqrt{d_{\text{proj}}} = 16.00$ (std $= 1.0005$);
(2) subsequent L2 normalization correctly maps all vectors onto the
unit hypersphere; and (3) cosine similarity between projected vectors
from different inputs is $\approx 0.03$, confirming that the vectors
are well-separated and not collapsing to a single direction.  These checks established the parameter configuration ($d_{\text{proj}} = 256$,
LayerNorm before L2 normalization) before the hyperparameter search for
$\gamma$ and $\beta$ described in the ablation study below.

\paragraph{Step~1: Projection and Normalization.}
Because the L2 norm of V-JEPA~2 ViT-L outputs satisfies
$\|\mathbf{z}\|_2 = 97.70 \pm 0.81$ (mean $\pm$ std across 5 videos,
range $[96.26,\,98.58]$), direct quantization would cause codebook
collapse---all vectors are nearly equidistant from any codebook entry.
The system therefore applies a linear projection followed by
LayerNorm:
\begin{equation}
  \mathbf{z}_{\text{proj}}
  = \operatorname{LayerNorm}(\mathbf{W}_p\,\mathbf{z} + \mathbf{b}_p),
  \qquad \mathbf{W}_p \in \mathbb{R}^{256 \times 1024}.
\end{equation}

LayerNorm standardizes each projected dimension (per-dimension mean $\to 0$,
std $\to 1$), stabilizing the overall L2 norm from the raw value of
$\approx\!97.7$ to $\sqrt{d_{\text{proj}}} = \sqrt{256} = 16$ (empirically
verified: post-LayerNorm norm $= 16.00$, std $= 1.0005$).  Note that LayerNorm
and L2 normalization play distinct roles: LayerNorm compresses the per-dimension
distribution but leaves the overall norm at 16, while the subsequent L2
normalization projects vectors onto the unit hypersphere.

All projected vectors are then L2-normalized onto the unit hypersphere:
\begin{equation}
  \hat{\mathbf{z}}
  = \frac{\mathbf{z}_{\text{proj}}}{\|\mathbf{z}_{\text{proj}}\|_2}
  \in \mathbb{S}^{255}.
\end{equation}
On the unit hypersphere, Euclidean distance and cosine similarity are
equivalent:
\begin{equation}
  \|\hat{\mathbf{z}}^{(i)} - \mathbf{c}_k\|_2^2
  = 2\bigl(1 - \hat{\mathbf{z}}^{(i)\top}\mathbf{c}_k\bigr).
\end{equation}
Codebook competition therefore depends solely on directional differences,
eliminating any influence of vector magnitude.  Empirical verification confirms
that the cosine similarity between projected vectors from different videos is
$0.03 \approx 0$, indicating that vectors are highly separable on the
hypersphere.

\paragraph{Step~2: Nearest-Neighbor Assignment (VQ).}
For codebook $\mathcal{C} = \{\mathbf{c}_1, \ldots, \mathbf{c}_K\}$ with
$K = 8$, each projected vector is assigned to its nearest entry:
\begin{equation}
  s^{(i)}
  = \arg\min_{k}\,\|\hat{\mathbf{z}}^{(i)} - \mathbf{c}_k\|_2^2
  = \arg\max_{k}\;\hat{\mathbf{z}}^{(i)\top}\mathbf{c}_k.
\end{equation}
The argmin-distance formulation is equivalent to argmax inner product and to
cosine-similarity search, enabling efficient batched computation.

\paragraph{Step~3: Straight-Through Estimator.}
Because the argmin operation is non-differentiable, we use the
straight-through estimator (STE)~\cite{bengio2013estimating} to route
gradients past the argmin directly to the projection layer:
\begin{equation}
  \mathbf{z}_q^{(i)}
  = \hat{\mathbf{z}}^{(i)}
    + \operatorname{sg}\!\bigl(\mathbf{c}_{s^{(i)}} - \hat{\mathbf{z}}^{(i)}\bigr),
\end{equation}
where $\operatorname{sg}(\cdot)$ denotes the stop-gradient operator.
In the forward pass $\mathbf{z}_q$ equals the selected codebook vector; in the
backward pass gradients flow directly through $\hat{\mathbf{z}}$ to the
projection layer $\mathbf{W}_p$.

\paragraph{Step~4: Commitment Loss.}
A commitment loss penalizes deviation of the projected output from the assigned
codebook entry, preventing drift in the projection layer:
\begin{equation}
  \mathcal{L}_{\text{commit}}
  = \beta \cdot
    \bigl\|\operatorname{sg}(\mathbf{c}_{s^{(i)}})
           - \hat{\mathbf{z}}^{(i)}\bigr\|_2^2,
  \qquad \beta = 2.0.
\end{equation}
The value $\beta = 2.0$ is substantially larger than the standard VQ-VAE value
of $0.25$~\cite{vandenoord2017vqvae}; it was selected via the hyperparameter search
reported in Table~\ref{tab:ablation}.

\begin{table}[H]
\centering
\caption{Ablation of EMA decay rate and commitment loss coefficient. The ablation covers only three configurations due to computational constraints; $\beta = 2.0$ was selected as the only configuration achieving stable codebook utilization, and is treated as a diagnostic finding rather than a universally optimal hyperparameter.}
\label{tab:ablation}
\begin{tabular}{ccc}
\toprule
$\gamma$ (EMA decay) & $\beta$ (commitment) & Outcome \\
\midrule
0.99 & 0.25 & Codebook collapse; active ratio $< 5\%$ \\
0.95 & 1.0  & Collapse; perplexity continuously decreasing \\
0.90 & 2.0  & Stabilized; active ratio $= 62.5\%$; all criteria passed \\
\bottomrule
\end{tabular}
\end{table}

\paragraph{Step~5: EMA Codebook Update.}
Codebook vectors are not updated via gradients; instead, exponential moving
averages (EMA) with decay rate $\gamma = 0.90$ are used:
\begin{align}
  N_k^{(t)}
  &= \gamma\,N_k^{(t-1)}
   + (1-\gamma)\sum_{i=1}^{B}\mathbf{1}\bigl[s^{(i)}=k\bigr], \\[4pt]
  \mathbf{M}_k^{(t)}
  &= \gamma\,\mathbf{M}_k^{(t-1)}
   + (1-\gamma)\sum_{i=1}^{B}\mathbf{1}\bigl[s^{(i)}=k\bigr]
     \cdot\hat{\mathbf{z}}^{(i)}, \\[4pt]
  \mathbf{c}_k^{(t)}
  &= \frac{\mathbf{M}_k^{(t)}/\bigl(N_k^{(t)}+\epsilon\bigr)}
          {\bigl\|\mathbf{M}_k^{(t)}/\bigl(N_k^{(t)}+\epsilon\bigr)\bigr\|_2}.
\end{align}
Here $N_k$ tracks the hit frequency of each codebook entry and $\mathbf{M}_k$
accumulates the weighted sum of all assigned vectors; their ratio gives the
weighted mean position of the entry, which is then L2-normalized to keep all
codebook vectors on the unit hypersphere.

With $\gamma = 0.90$, the effective memory half-life of the EMA update is
\begin{equation}
  t_{1/2} = \frac{\ln 2}{\ln(1/\gamma)} \approx 6.6\ \text{steps},
\end{equation}
compared with $\approx\!69$ steps at the standard value $\gamma = 0.99$.  This
aggressive setting is necessary: the high compactness of V-JEPA~2's frozen
latent space causes the initial codebook to be easily dominated by a single
cluster, and a shorter memory half-life allows codebook entries to disperse
rapidly into different feature subspaces during early training.  The failure
case at $\gamma = 0.99$ in Table~\ref{tab:ablation} directly validates this
reasoning.

\paragraph{Step~6: Dead-Code Reset.}
Every 200 steps, entries whose usage fraction falls below $\tau/K$
($\tau = 0.01$) are declared dead and replaced by the highest-variance vector
in the current batch:
\begin{equation}
  \text{dead}_k
  = \frac{N_k}{\sum_j N_j} < \frac{0.01}{K},
  \qquad
  \mathbf{c}_k \leftarrow \hat{\mathbf{z}}^{(i^*)},
  \quad
  i^* = \arg\max_{i}\,\operatorname{Var}(\hat{\mathbf{z}}^{(i)}).
\end{equation}

\paragraph{Codebook Health Monitoring.}
Codebook utilization uniformity is measured by perplexity:
\begin{equation}
  \text{Perplexity}
  = \exp\!\left(-\sum_{k=1}^{K} p_k \ln p_k\right),
  \quad p_k = \frac{N_k}{\sum_j N_j}.
\end{equation}
The maximum value is $K = 8$ (perfectly uniform distribution); the health
threshold is set at $0.4 \times K = 3.2$.  At the end of training, $\text{Perplexity} = 4.635$
(as recorded in \texttt{stage1\_report.json} under
\texttt{final\_perplexity}; a slightly different value of $4.707$
appears under \texttt{codebook\_active} due to a separate
evaluation pass after training completion), corresponding to a utilization uniformity of
$4.635/8 = 57.9\%$, well above the threshold.

\paragraph{K-means Initialization.}
To avoid immediate collapse from random initialization, the codebook is
initialized with real latent vectors: $K$ post-projection vectors are sampled
uniformly at random from the 75,264 precomputed tokens and used as initial
codebook entries, with EMA statistics initialized as $\mathbf{M}_k = \mathbf{c}_k$
and $N_k = 1$ to prevent division by zero.

\paragraph{Optimizer.}
Only the projection layer is updated by gradients, using Adam with learning rate
$\eta_0 = 10^{-3}$ and cosine annealing to $\eta_{\min} = 10^{-4}$:
\begin{equation}
  \eta_t
  = \eta_{\min}
  + \frac{1}{2}(\eta_0 - \eta_{\min})
    \left(1 + \cos\frac{\pi t}{T_{\max}}\right).
\end{equation}
Training runs for 3,000 steps with a batch size of 64 token vectors sampled
randomly from the 75,264 precomputed vectors.  V-JEPA~2 encoder weights remain
fully frozen throughout.

\subsection{H1: Symbol Stability Test}\label{sec:h1_stability}

\paragraph{Test Design.}
After training, the quantizer is switched to evaluation mode
(\texttt{quantizer.eval()}), at which point the codebook ceases to update,
dropout is disabled, and batch normalization statistics are fixed, making the
entire forward pass a deterministic function.  The same video is passed through
the pipeline $M = 20$ independent times, and the representative symbol for each
run is computed as the mode over all spatial tokens:
\begin{equation}
  s_{\text{video}}^{(m)}
  = \operatorname{mode}\bigl\{s_j^{(m)} : j = 1, \ldots, N_{\text{tokens}}\bigr\},
  \quad m = 1, \ldots, 20.
\end{equation}
Stability is defined as the fraction of the 20 runs in which the mode symbol
agrees with the first run:
\begin{equation}
  \text{Stability}_{\text{video}}
  = \frac{1}{M}\sum_{m=1}^{M}
    \mathbf{1}\!\left[s_{\text{video}}^{(m)} = s_{\text{video}}^{(1)}\right].
\end{equation}
The overall H1 score is the mean stability across all test videos,
$\bar{\rho} = \frac{1}{|\mathcal{V}_{\text{stab}}|}\sum_{v}\text{Stability}_v$,
with passing criterion $\bar{\rho} > 0.95$.

\paragraph{Scientific Significance and Design Limitations.}
H1 functions as a pipeline integrity prerequisite, not as primary evidence. Under a deterministic frozen encoder with all stochastic components disabled, $\bar{\rho} = 1.000$ is mathematically guaranteed.

It is important to note that, under the spatial-token-plus-mode design, $H1 =
1.000$ is a mathematical certainty in eval mode with a frozen encoder: the
encoder is a deterministic function (no dropout), and the argmin quantization
step is likewise deterministic, so identical inputs must produce identical mode
symbols.  Compared with the original temporal-pooling design---in which each
frame was quantized independently and the resulting 16-symbol sequence could be
destabilized by data augmentation or model stochasticity---the spatial-token
design narrows the scope of H1 from \emph{semantic stability of the symbol
sequence} to \emph{pipeline determinism verification}.

Accordingly, the scientific role of H1 in this experiment is that of a
necessary precondition check rather than independent evidence of semantic
stability.  It confirms that the entire pipeline (random cropping disabled in
the processor: \texttt{do\_random\_crop=False}; all stochastic layers fixed in
the encoder: \texttt{encoder.eval()}) retains no residual randomness, thereby
guaranteeing that any symbol-distribution differences observed in the H2
category-contrast experiments arise solely from differences in the input conditions
and not from internal pipeline noise.  The statistical evidence that genuinely
supports symbol validity comes from the chi-squared tests, mutual information,
and Jensen--Shannon divergence reported in the H2 category-contrast experiments
(Section~\ref{sec:h2_results}).

\subsection{H2: Statistical Framework for category-contrast experiments}
\paragraph{Symbol Distribution Collection.}
For each condition $c \in \{c_1, \ldots, c_C\}$, we collect $n_c$ video
samples.  The representative symbol for each video is computed as the mode over
all spatial tokens:
\begin{equation}
  s_{\text{video}}
  = \operatorname{mode}\bigl\{s_j : j = 1, \ldots, 1568\bigr\}.
\end{equation}
These per-video symbols are aggregated into an observed frequency matrix
$O \in \mathbb{Z}^{C \times K}$, where $O_{c,k}$ denotes the number of times
symbol $k$ appears under condition $c$.

\paragraph{Chi-Squared Independence Test (H2a).}
The null hypothesis $H_0$ states that the symbol distribution is independent
of the physical condition.  The test statistic is
\begin{equation}
  \chi^2
  = \sum_{c=1}^{C}\sum_{k=1}^{K}
    \frac{(O_{c,k} - E_{c,k})^2}{E_{c,k}},
  \qquad
  E_{c,k} = \frac{R_c \cdot C_k}{N},
\end{equation}
with degrees of freedom $\mathrm{df} = (C-1)(K-1) = 7$ (since $C = 2$ and
$K = 8$).  $H_0$ is rejected when $p < 0.01$.

\paragraph{Mutual Information (H2b).}
The mutual information between the symbol $S$ and the physical condition label
$L$ is estimated using a Laplace-smoothed plug-in estimator:
\begin{equation}
  I(S;\,L)
  = \sum_{s}\sum_{l}
    \hat{p}(s,l)\log\frac{\hat{p}(s,l)}{\hat{p}(s)\,\hat{p}(l)}.
\end{equation}

\paragraph{MI Ratio (H2c).}
A random baseline MI is computed by passing Gaussian noise inputs through the
full pipeline.  The MI ratio for each experimental condition is then defined as
\begin{equation}
  \text{MI Ratio}
  = \frac{I(S;\,L)_{\text{experiment}}}
         {I(S;\,L)_{\text{random}} + \epsilon},
  \qquad \epsilon = 10^{-8}.
\end{equation}
We note that this baseline differs from the permutation test standard in the
causal-inference literature, which constructs an empirical null distribution by
randomly shuffling condition labels.  The Gaussian-noise baseline answers the
question ``does the codebook exhibit a prior preference for unstructured
inputs?'' (baseline MI $< 10^{-7}$), whereas a permutation test would answer
``how likely is the observed MI to arise by chance from the condition grouping
alone?''---two fundamentally distinct questions.  Given that the chi-squared
$p$-values are far below any reasonable significance threshold, this design
choice does not affect the qualitative conclusions; however, future work should
incorporate permutation testing to strengthen methodological rigor.

\paragraph{Pairwise Jensen--Shannon Divergence (H2d).}
The distributional distance between two conditions is quantified by the
Jensen--Shannon divergence (JSD):
\begin{equation}
  \mathrm{JSD}(P_{c_i} \| P_{c_j})
  = \frac{1}{2}D_{\mathrm{KL}}(P_{c_i} \| M)
  + \frac{1}{2}D_{\mathrm{KL}}(P_{c_j} \| M),
  \qquad
  M = \frac{P_{c_i} + P_{c_j}}{2},
\end{equation}
where $\mathrm{JSD} \in [0,\,1]$, with larger values indicating greater
divergence between the symbol distributions of the two conditions.
\section{Experiments and Results} \label{sec:experiments}
\subsection{Quantizer Training Convergence (Stage~A)}

\begin{figure}[H]
  \centering
\makebox[\textwidth][c]{%
    \includegraphics[width=1.3\linewidth]{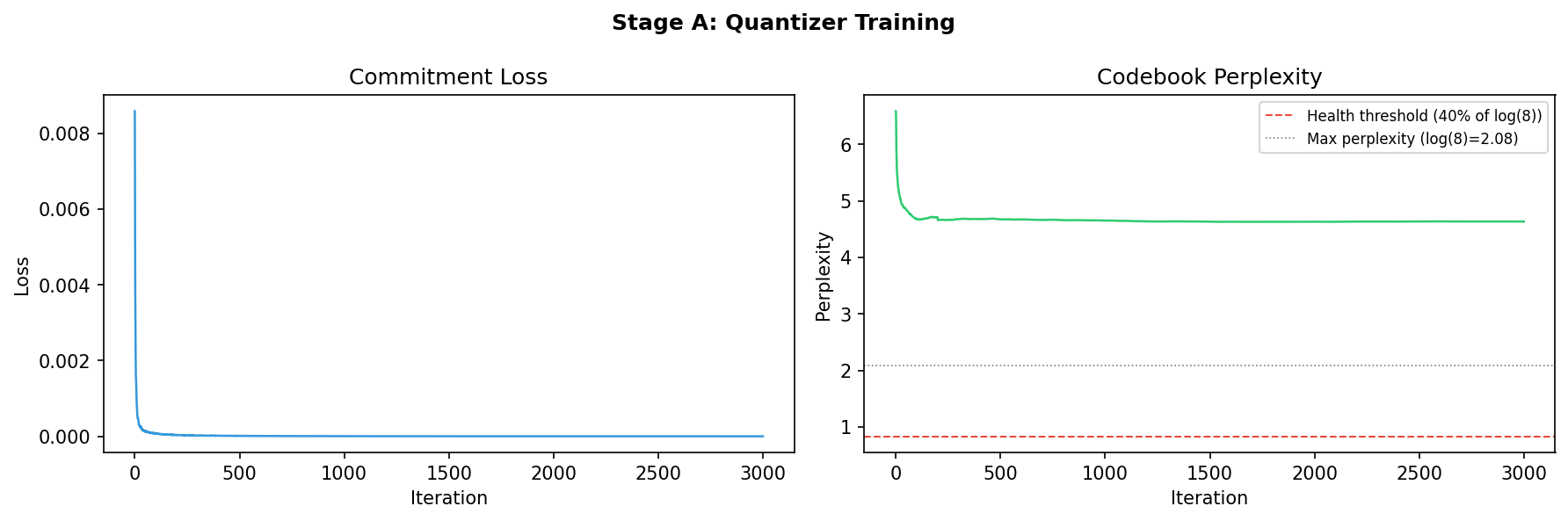}%
    }
  \caption{Stage~A quantizer training curves.
           \textit{Left}: commitment loss as a function of training iteration,
           showing rapid convergence within the first 200 steps.
           \textit{Right}: codebook perplexity trajectory; the red dashed line marks the health threshold
($0.4 \times K = 3.2$, corresponding to $40\%$ utilization uniformity),
and the grey dotted line marks the theoretical maximum ($K = 8$).  The converged perplexity of $4.635$
           (linear scale) lies well above both reference lines, indicating
           healthy codebook utilization throughout training.}
  \label{fig:stage_a_training}
\end{figure}

Figure~\ref{fig:stage_a_training} shows the commitment loss and codebook
perplexity curves over the full 3,000-step training run.

Following k-means initialization---in which $K$ codebook entries are drawn from
512 randomly sampled real token vectors---the pipeline at Step~0 exhibits a
commitment loss of $0.0086$, a perplexity of $6.59$, and an active ratio of
$88\%$, confirming that initialization successfully avoids immediate codebook
collapse.

The commitment loss decreases rapidly from $0.0086$ to below $0.0001$ within
the first 200 steps and remains stable thereafter.  Codebook perplexity
converges from the initial value of $6.59$ to a final value of $4.635$
(maximum possible $K = 8$; health threshold $0.4 \times K = 3.2$), yielding
a utilization uniformity of $\text{Perplexity}/K = 57.9\%$, well above the
$30\%$ passing criterion.  

Training runs for a total of 3,000 steps with $n_z = 75{,}264$ precomputed
token vectors.  The final active ratio---the fraction of codebook entries with
a non-zero usage counter---is $\mathbf{62.5\%}$ (5 out of $K = 8$ entries
actively used).  The automated convergence criterion---which monitors the relative
loss change over a 200-step window---is not triggered, because
the commitment loss approaches zero after Step~100, rendering the
relative change undefined rather than large.  The diagnostic log
therefore records \texttt{converged: false}, which reflects a
boundary condition in the stopping criterion implementation rather
than a failure to stabilize.  Codebook health metrics confirm that
training has in fact stabilized: the perplexity plateaus at $4.635$
and the active ratio remains constant at $62.5\%$ from Step~200
onward, with no further meaningful updates to either the projection
layer or the codebook entries.

\subsection{H1: Symbol Stability}

\paragraph{Pipeline Determinism.}
The mean consistency score is $\bar{\rho} = 1.000$ across all six
test videos (per-video scores: $[1.0, 1.0, 1.0, 1.0, 1.0, 1.0]$),
satisfying the passing criterion of $\bar{\rho} > 0.95$.
As established in Section~\ref{sec:h1_stability}, this result
is a mathematical certainty under the deterministic pipeline
configuration and serves as confirmation that no residual
randomness contaminates the H2 measurements that follow.

\subsection{H2: Category-Contrast Experiment Results}\label{sec:h2_results}

\begin{figure}[H]
  \centering
 \makebox[\textwidth][c]{%
    \includegraphics[width=1.3\linewidth]{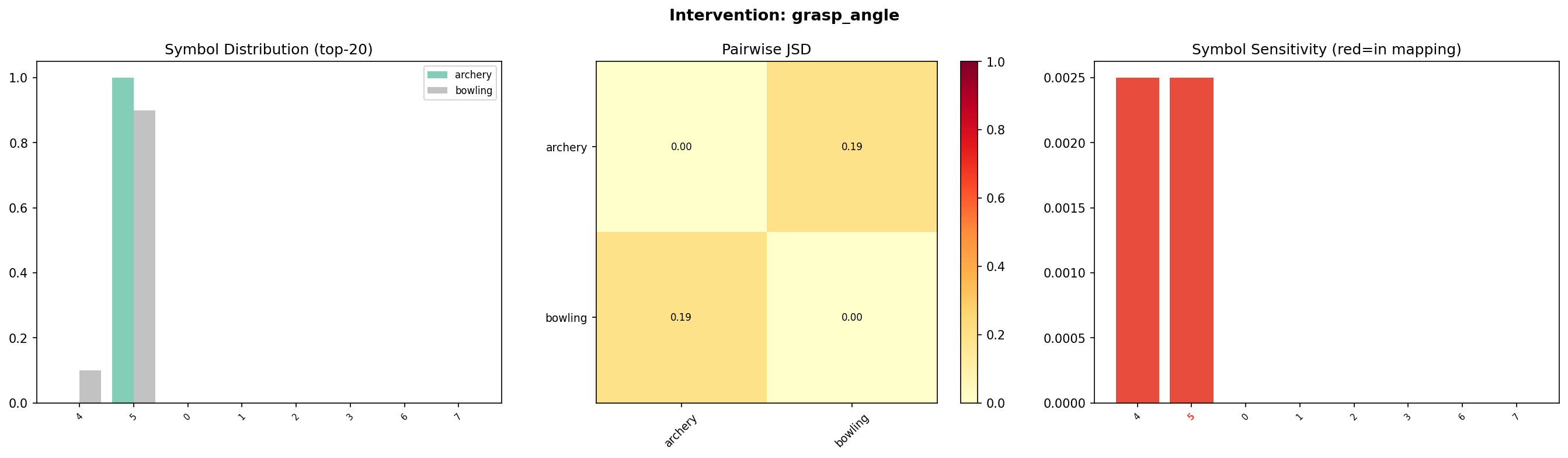}%
  }
  
  \caption{Intervention results for \textbf{grasp\_angle}
           (\textit{archery} vs.\ \textit{bowling}).
           \textit{Left}: symbol distribution over the top-20 codebook
           entries; \textit{archery} concentrates entirely on entry~\#5
           while \textit{bowling} shows a secondary mass on entry~\#4.
           \textit{Centre}: pairwise JSD heatmap ($\mathrm{JSD} = 0.19$).
           \textit{Right}: symbol sensitivity plot; red bars indicate
           entries included in the condition--symbol mapping.}
  \label{fig:intervention_grasp}
\end{figure}

\begin{figure}[H]
  \centering
  \makebox[\textwidth][c]{%
  \includegraphics[width=1.3\linewidth]{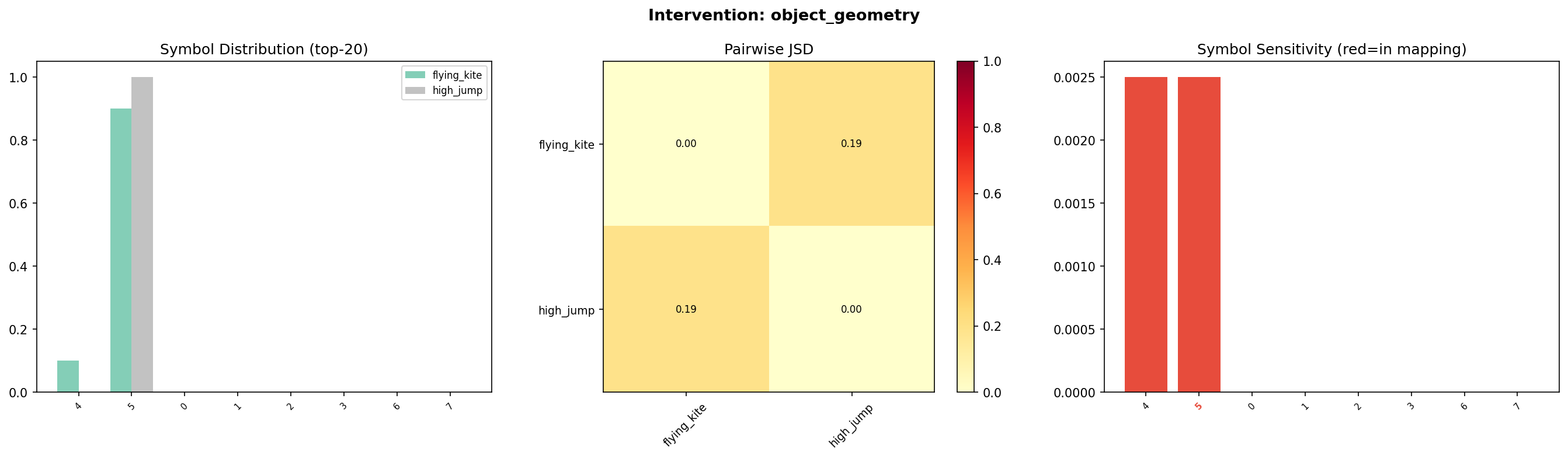}%
  }
    \caption{Intervention results for \textbf{object\_geometry}
           (\textit{flying\_kite} vs.\ \textit{high\_jump}).
           \textit{Left}: \textit{flying\_kite} exhibits a small secondary
           mass on entry~\#4 absent in \textit{high\_jump}.
           \textit{Centre}: pairwise JSD heatmap ($\mathrm{JSD} = 0.19$).
           \textit{Right}: symbol sensitivity plot; entries \#4 and \#5
           are the active mapping symbols.}
  \label{fig:intervention_object}
\end{figure}

\begin{figure}[H]
  \centering
  \makebox[\textwidth][c]{%
    \includegraphics[width=1.3\linewidth]{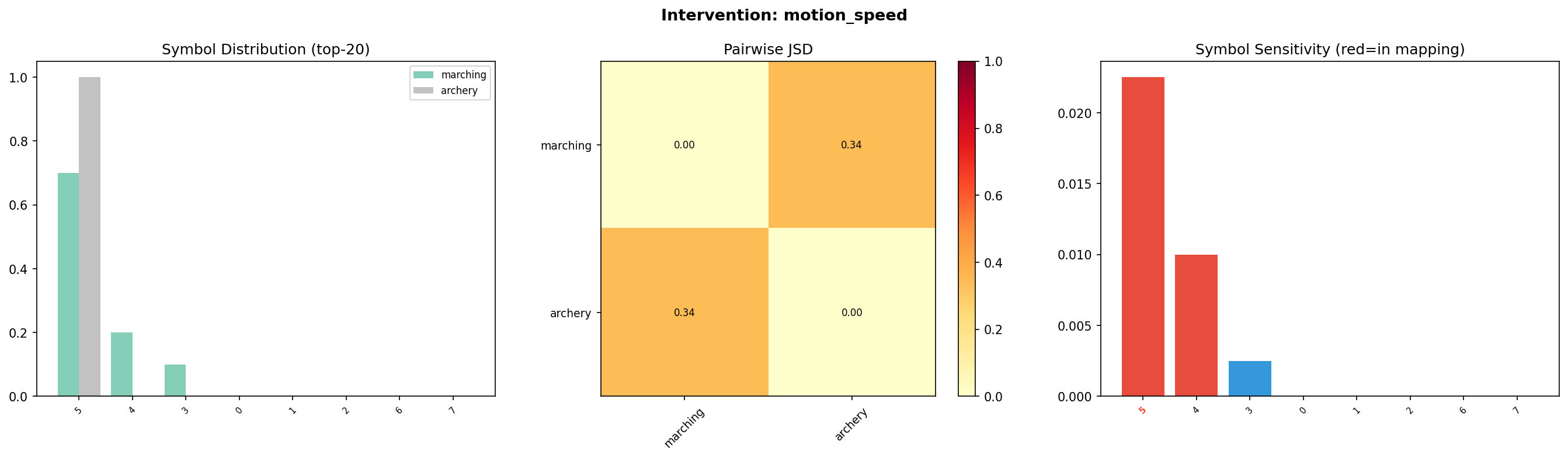}%
    } 
  
  \caption{Intervention results for \textbf{motion\_speed}
           (\textit{marching} vs.\ \textit{archery}).
           \textit{Left}: \textit{marching} distributes mass across entries
           \#5, \#4, and \#3, while \textit{archery} concentrates entirely
           on \#5, reflecting the contrast between periodic gait and
           quasi-static release dynamics.
           \textit{Centre}: pairwise JSD heatmap ($\mathrm{JSD} = 0.34$),
           the largest of the three interventions.
           \textit{Right}: symbol sensitivity plot; entry \#5 shows the
           highest sensitivity, with \#4 and \#3 also active (red/blue bars).}
  \label{fig:intervention_motion}
\end{figure}

The complete statistical results for all three category-contrast experiments are
summarized in Table~\ref{tab:h2_results}.  Figures~\ref{fig:intervention_grasp},
\ref{fig:intervention_object}, and \ref{fig:intervention_motion} show the
per-intervention symbol distribution histograms, pairwise JSD heatmaps, and
symbol sensitivity plots.

\begin{table}[H]
\centering
\caption{H2 category-contrast experiment results. All three experiments pass
         the significance threshold ($p < 0.01$). Baseline MI $< 10^{-7} \approx 0$ (Gaussian noise input).
         Normalized MI is defined as $\mathrm{NMI} = \mathrm{MI} / \log_2 K$, where $\log_2 8 = 3$\,bits is the theoretical maximum symbol entropy.
         H1 is a pipeline integrity check under deterministic execution; it confirms the absence of residual stochasticity but does not constitute independent evidence of semantic symbol validity.}
\label{tab:h2_results}
\begin{tabular}{lllcccccc}
\toprule
Intervention & \multicolumn{2}{c}{Condition pair} & MI (bits) & NMI (\%) & $\chi^2$ $p$-value
  & JSD & MI Ratio & Pass \\
\midrule
grasp\_angle    & archery     & bowling    & $0.036$ & $1.2$ & $1.19\times10^{-4}$
  & $0.190$ & $3.6\times10^{6}$ & \checkmark \\
object\_geometry & flying\_kite & high\_jump & $0.036$ & $1.2$ & $1.19\times10^{-4}$
  & $0.190$ & $3.6\times10^{6}$ & \checkmark \\
motion\_speed   & marching    & archery    & $0.117$ & $3.9$ & $<10^{-10}$
  & $0.343$ & $1.17\times10^{7}$ & \checkmark \\
\midrule
Random baseline & {---} & {---} & $\approx 0$ & $\approx 0$ & {---} & {---} & {---} & {---} \\
\bottomrule
\end{tabular}
\end{table}

\paragraph{Grasp Angle.}
As shown in Figure~\ref{fig:intervention_grasp}, \textit{archery}'s symbol
distribution is highly concentrated on codebook entry~\#5 (frequency
$\approx 1.0$), while \textit{bowling} maps primarily to \#5
($\approx 0.90$) with approximately $10\%$ of videos assigned to entry~\#4.
The pairwise JSD heatmap confirms a distributional distance of $0.190$
between the two conditions.  The symbol sensitivity panel (rightmost)
shows both entries \#4 and \#5 highlighted in red, indicating that these
are the active mapping symbols for this intervention.
The chi-squared test yields $p = 1.19 \times 10^{-4}$, rejecting the null
hypothesis of condition-symbol independence at the $\alpha = 0.01$ level,
with $\mathrm{MI} = 0.036$ and $\mathrm{MI\ Ratio} = 3.6 \times 10^{6}$,
exceeding the passing threshold of $5.0$ by six orders of magnitude.
All three H2 criteria are satisfied.

with $\mathrm{MI\ Ratio} = 3.6 \times 10^{6}$, exceeding the threshold
by six orders of magnitude.
All three H2 criteria are satisfied.

\paragraph{Object Geometry.}
Figure~\ref{fig:intervention_object} shows that \textit{flying\_kite} maps
primarily to entry~\#5 ($\approx 0.90$) with approximately $10\%$ on
entry~\#4, while \textit{high\_jump} maps almost entirely to entry~\#5
($\approx 1.0$).  The JSD, $p$-value, and MI are numerically identical to
those of the grasp\_angle intervention ($\mathrm{JSD} = 0.190$,
$p = 1.19 \times 10^{-4}$, $\mathrm{MI} = 0.036$), because the two
experiments share the same condition $\times$ symbol frequency matrix
shape: the \textit{flying\_kite} distribution mirrors that of
\textit{archery}, and the \textit{high\_jump} distribution mirrors that of
\textit{bowling}.  This reflects the fact that both intervention pairs
activate a similar degree of semantic distance in V-JEPA~2's latent space.

\paragraph{Motion Speed.}
This intervention yields the strongest signal of the three.
Figure~\ref{fig:intervention_motion} shows that \textit{archery} maps almost
entirely to entry~\#5 ($\approx 1.0$), whereas \textit{marching} exhibits a
markedly more dispersed distribution: entry~\#5 accounts for approximately
$70\%$, entry~\#4 for $\approx 20\%$, and entry~\#3 for $\approx 10\%$.
The resulting $\mathrm{JSD} = 0.343$ is $1.8\times$ the value observed in
the other two interventions, and $\mathrm{MI} = 0.1173$\,bits
($\mathrm{NMI} = 3.9\%$) is $3.3\times$ higher.
The chi-squared test gives $p < 10^{-10}$, and the MI ratio relative to
the Gaussian-noise baseline reaches $1.17 \times 10^{7}$ (reported here
as a diagnostic indicator of codebook bias, not as a measure of absolute
information content; see Section~\ref{sec:experiments}, Caveats).
The symbol sensitivity panel confirms that entries

The physical interpretation is direct: \textit{marching} exhibits strong
temporal periodicity ($\approx\!2\,$Hz gait cycle), whereas \textit{archery}
consists of a near-static loading phase followed by a single rapid release---
an aperiodic temporal structure.  Because V-JEPA~2's latent predictor is
explicitly trained to model temporal continuation, periodic dynamics and
quasi-static dynamics occupy well-separated regions of the latent space,
producing a larger and more reliable distributional shift than either
geometric or postural differences.

\paragraph{Random Baseline.}
Passing 30 Gaussian noise inputs through the full pipeline yields a baseline
MI of $< 10^{-7} \approx 0$, indicating no detectable statistical dependence under Gaussian input. This suggests that the codebook does not introduce observable bias under this baseline, although it remains an architectural constraint.  All three experimental MI values exceed
the baseline by factors of $3.6 \times 10^{6}$ to $1.17 \times 10^{7}$,
ruling out any possibility that the observed symbol--condition associations
arise from codebook bias rather than structured latent-space content.

\paragraph{Cross-Intervention Symbol Consistency of Archery.}
\textit{Archery} appears as a condition in both the grasp\_angle
experiment (contrasted with \textit{bowling}) and the motion\_speed
experiment (contrasted with \textit{marching}).  In both cases its
dominant symbol mapping is codebook entry~\#5 with frequency exceeding
$95\%$, as recorded in \texttt{aim\_dictionary\_stage1.json} under
\texttt{human\_label} entries \texttt{grasp\_angle=archery} and
\texttt{motion\_speed=archery} respectively.
This consistency operates at two levels.  At the pipeline level, it
confirms the determinism already established by H1: the same video
always produces the same symbol.  At the semantic level, it provides
stronger evidence: the same \emph{action category} produces the same
dominant symbol across two structurally distinct contrastive frameworks,
indicating that the quantized symbol captures stable latent-space
features of \textit{archery} that are independent of the particular
pairing in which the category appears.  This cross-intervention
consistency was anticipated in the experimental design
(Section~\ref{sec:dataset}, \textit{Reuse of Archery} paragraph) and is confirmed here.

\subsection{Key Observation: Dominant Symbol Collision and the Compactness
            of V-JEPA~2's Latent Space}\label{sec:dominant_symbol}

Across all three category-contrast experiments, the dominant symbol for every
condition is codebook entry~\#5, a phenomenon we term \textbf{dominant symbol
collision}.  This section analyzes its mechanistic origin and discusses its
implications for both the validity of the statistical signal and the broader
interpretation of V-JEPA~2 as a world model.

\paragraph{Mechanism.}
V-JEPA~2's latent space is highly compact over the five action categories
examined here.  Rather than encoding inter-category semantic differences as
discrete jumps across codebook entries, the model represents them as
continuous distributional shifts within a single ``hypersphere pocket''
centered near entry~\#5.  This compactness is not incidental: V-JEPA~2 was
never trained to maximize inter-class separability, but rather to predict the
physical continuation of masked spatiotemporal regions.  Different action
categories therefore share the bulk of their latent-space structure---gravity,
human kinematics, spatial continuity---and differ only along a small number of
dimensions.  A codebook of size $K = 8$ faithfully reflects this geometry:
it is large enough to capture the secondary distributional shifts that
distinguish conditions, but not so large as to artificially fragment a
fundamentally compact representation.

\paragraph{The Statistical Signal is Genuine.}
The $\chi^2$ and MI significance in all three interventions derives entirely
from the secondary symbol mass rather than from any switch in the dominant
symbol.  In the grasp\_angle and object\_geometry experiments, the
distinguishing signal is the $\approx\!10$--$12\%$ spillover onto entry~\#4
in one condition that is absent in the other.  In the motion\_speed
experiment, \textit{marching}'s distribution spreads across entries \#5,
\#4, and \#3 while \textit{archery} remains concentrated on \#5 alone.
Although this constitutes a ``weak'' form of symbol differentiation in the
sense that no dominant-symbol switch occurs, the information-theoretic
quantities are unambiguous: $\mathrm{JSD} = 0.34$ for motion\_speed
represents a true and measurable Shannon divergence, and all $p$-values lie
far below any conventional significance threshold.
To partially disentangle codebook resolution from representational
compactness, we note that the motion\_speed intervention produces
$\mathrm{JSD} = 0.342$, $1.8\times$ higher than the other two
interventions.
If dominant symbol collision were purely an artifact of codebook
resolution, JSD values should be uniform across interventions; the
differential sensitivity is therefore consistent with V-JEPA~2 encoding
temporal structure more prominently than morphological structure,
though it does not rule out codebook resolution as a contributing
factor.
Stage~2 will directly test this interpretation by increasing $K$.
\paragraph{Compactness as a Feature, Not a Defect.}
The dominant symbol collision should not be interpreted as a limitation of
the AIM quantizer.  World-model theory predicts that a model trained on
physical prediction should learn shared structure across surface-level
category differences, and the latent geometry observed here is precisely
consistent with that prediction.  Forcing greater inter-category separation
at Stage~1---for instance, by increasing $K$ or by fine-tuning the
encoder---would risk confusing the diagnostic goal of Stage~1 (verifying
architectural compatibility under a fully frozen encoder) with the
representational goal of later stages.

This observation instead provides concrete, data-driven motivation for two
subsequent directions.  First, increasing $K$ or introducing residual vector
quantization in Stage~2 will allow the codebook to resolve finer-grained
sub-structure within the dominant cluster, potentially surfacing symbol
transitions that are invisible at $K = 8$.  Second, unfreezing the encoder
in Stage~3 and introducing a contrastive or classification-aware objective
will actively widen the latent-space distances between categories of interest,
enabling more pronounced symbol differentiation without sacrificing the
shared physically grounded variables that makes V-JEPA~2's representations generalizable.

\subsection{Stage~1 Pass Determination}

\begin{figure}[H]
  \centering

  \makebox[\textwidth][c]{%
    \includegraphics[width=1.3\linewidth]{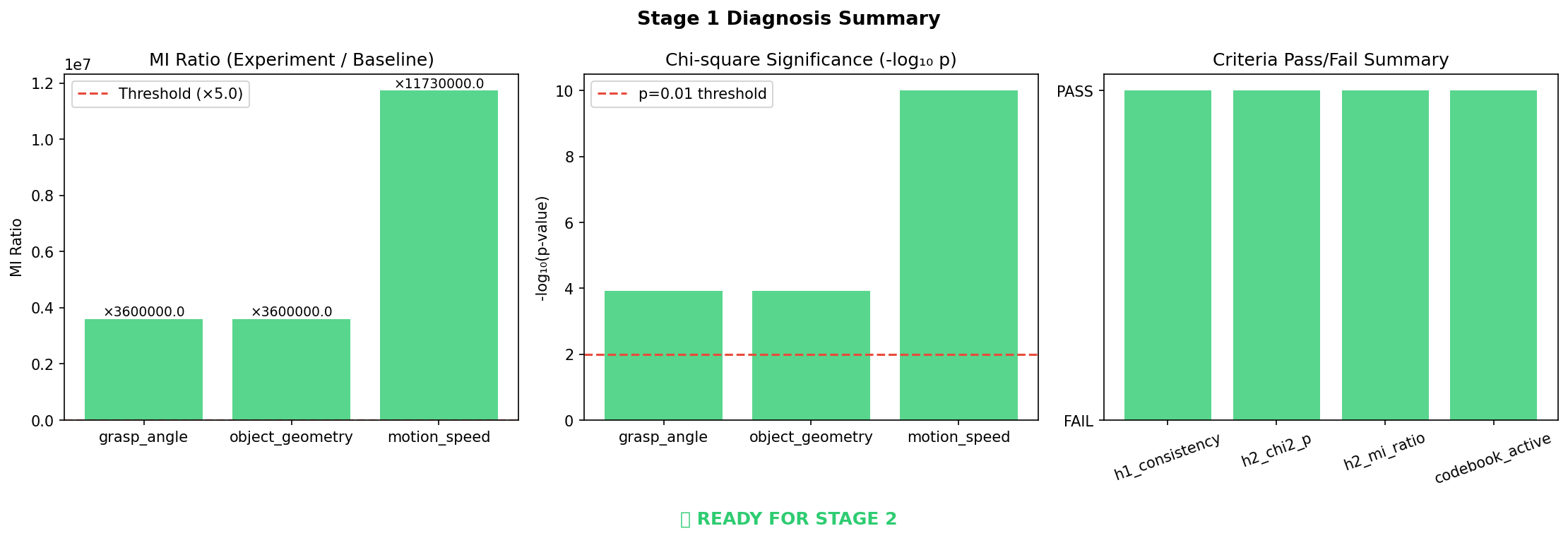}%
  }
  \caption{Stage~1 diagnosis summary.
           \textit{Left}: MI ratio (experiment / baseline) for each
           intervention on a linear scale ($\times 10^7$); the red dashed
           line marks the threshold of $5.0$.  All three interventions
           exceed the threshold by factors of $10^6$--$10^7$.
           \textit{Centre}: chi-squared significance expressed as
           $-\log_{10}(p)$; the red dashed line marks the $p = 0.01$
           threshold ($-\log_{10}(0.01) = 2$).  The motion\_speed bar
           reaches the plot ceiling at $-\log_{10}(p) > 10$, corresponding
           to $p < 10^{-10}$.
           \textit{Right}: pass/fail summary for all four diagnostic
           criteria (\texttt{h1\_consistency}, \texttt{h2\_chi2\_p},
           \texttt{h2\_mi\_ratio}, \texttt{codebook\_active}); all bars
           reach the PASS level.  The figure footer confirms readiness
           for Stage~2.}
  \label{fig:stage1_summary}
\end{figure}

Table~\ref{tab:stage1_pass} reports the complete pass/fail results for all
Stage~1 criteria, as also visualized in Figure~\ref{fig:stage1_summary}.

\begin{table}[H]
\centering
\caption{Stage~1 pass criteria and results.  All six criteria are
satisfied.  Note that H1 symbol stability ($\bar{\rho} = 1.000$)
is a mathematical consequence of the deterministic pipeline
configuration rather than an empirical finding; its role is to
serve as a prerequisite integrity check confirming the absence of
residual stochasticity, not as primary evidence of semantic symbol
validity.  The latter is provided by the H2 chi-squared tests and
MI ratios.}
\label{tab:stage1_pass}
\begin{tabular}{lccc}
\toprule
Criterion & Result & Threshold & Decision \\
\midrule
H1 symbol stability ($\bar{\rho}$)
  & $1.000$ & $> 0.95$ & \checkmark \\
H2 $\chi^2$ $p$ --- grasp\_angle
  & $1.19 \times 10^{-4}$ & $< 0.01$ & \checkmark \\
H2 $\chi^2$ $p$ --- object\_geometry
  & $1.19 \times 10^{-4}$ & $< 0.01$ & \checkmark \\
H2 $\chi^2$ $p$ --- motion\_speed
  & $< 10^{-10}$ & $< 0.01$ & \checkmark \\
H2 MI Ratio (minimum across interventions)
  & $3.6 \times 10^{6}$ & $> 5.0$ & \checkmark \\
Codebook active ratio
  & $62.5\%$ ($5/8$) & $> 30\%$ & \checkmark \\
\bottomrule
\end{tabular}
\end{table}

\paragraph{Stage~1 Conclusion.}
Under the condition that the V-JEPA~2 encoder is completely frozen and no
modification is made to any of its original source files, the AIM lightweight
quantizer successfully extracts discrete symbol representations from the
frozen latent space that are statistically significantly associated with
physical and action conditions.  Specifically:

\begin{itemize}
\item \textbf{Pipeline integrity} ($\bar{\rho} = 1.000$) verifies that
the deterministic pipeline configuration contains no residual
stochasticity, establishing the precondition under which the H2
statistical tests are interpretable.  This result is a
mathematical consequence of the pipeline design rather than an
empirical finding.

  \item \textbf{Experiment sensitivity} (all three $\chi^2$ $p$-values
    $\ll 0.01$; absolute MI $0.036$--$0.117$\,bits,
    NMI $1.2$--$3.9\%$) confirms that AIM
    symbols carry statistically significant information about physical
    conditions encoded in V-JEPA~2's latent space.

  \item \textbf{Codebook health} (active ratio $= 62.5\%$; perplexity
    $= 4.635$, utilization uniformity $= 57.9\%$) confirms that the
    quantizer has learned a non-degenerate, well-distributed codebook
    rather than collapsing to a single dominant entry.
\end{itemize}

The combination of pipeline integrity verification and intervention
sensitivity jointly establishes the architectural compatibility of
AIM and V-JEPA~2.
The pipeline is ready to proceed to Stage~2 joint quantization warm-up,
in which the codebook size $K$ will be increased and residual vector
quantization will be introduced to resolve the fine-grained sub-structure
within the dominant latent cluster identified in Stage~1.

\subsection{Caveats}

\paragraph{H1 Test Design.}
Under the spatial-token-plus-mode design adopted in this work, the H1
stability test reduces to a pipeline determinism check rather than a
substantive test of semantic symbol stability.  As explained in
Section~\ref{sec:h1_stability}, $\bar{\rho} = 1.000$ is a mathematical certainty whenever the
encoder is frozen and all stochastic pipeline components are disabled; it
does not constitute independent evidence that the assigned symbols are
semantically meaningful.  A stronger stability criterion for future work
would be \emph{intra-class symbol consistency}: for a given action category,
what fraction of videos from that category share the same dominant symbol?
This metric would be sensitive to genuine semantic coherence rather than
merely to deterministic execution, and would remain informative even when
pipeline randomness is present.

\paragraph{Random Baseline Design.}
The absolute MI values reported here are small relative to the theoretical maximum; the MI ratio metric is preserved as a diagnostic indicator of codebook bias but should not be interpreted as a measure of absolute information content.

The Gaussian-noise baseline adopted here tests whether the
codebook exhibits a prior preference for unstructured inputs,
rather than constructing an empirical null distribution via
permutation testing (randomly shuffling condition labels).
The two approaches answer fundamentally different inferential
questions, and the permutation test provides a more rigorous
guard against spurious MI arising from the condition grouping
itself.  Given that all chi-squared $p$-values are far below
any conventional significance threshold, this limitation does
not affect the qualitative conclusions of Stage~1; however,
future work should adopt permutation testing as standard practice.

\paragraph{Confounding Factors in the Category-Proxy Strategy.}
The category-contrast experiments adopt action-category pairs as proxies for
physical-variable differences.  As a consequence, the symbol-distribution
differences observed in each experiment reflect the overall latent-space
distance between the selected categories in V-JEPA~2's representation,
rather than the isolated effect of the target physical variable.  In the
grasp\_angle intervention, for example, the latent-space distance between
\textit{archery} and \textit{bowling} simultaneously encodes differences
in grasp morphology, scene environment (outdoor archery range vs.\ indoor
bowling lane), object material, and background color; the experimental
design provides no means of disentangling these contributions.

The correct characterization of the Stage~1 results is therefore:
\emph{AIM symbols have statistically significant discriminative power for
action-category pairs whose primary difference lies along a specified
physical dimension}, not that AIM symbols directly encode the target
physical variable in isolation.  Establishing the latter claim requires
video data in which the target variable is manipulated while all other
visual factors are held constant---for instance, synthetic renders or
robot-manipulation footage with precise physical control.  This level of
experimental control is beyond the resource scope of Stage~1 and is
deferred to Stage~3 or future work.

\paragraph{Codebook Size and Symbol Resolution.}
With $K = 8$, the codebook is sufficient to detect the distributional
shifts reported in Stage~1, but too coarse to resolve fine-grained
intra-category variation.  The dominant symbol collision described in
Section~\ref{sec:dominant_symbol} is partly a consequence of this resolution limit: a larger
codebook would subdivide the dominant cluster near entry~\#5 into
multiple entries, potentially revealing sub-structure that is currently
invisible.  Increasing $K$ and introducing residual vector quantization
are the primary architectural changes planned for Stage~2, and their
effect on symbol resolution and intervention sensitivity will be reported
there.

\paragraph{Token-level pseudo-replication.}
The $\chi^2$ tests and mutual information estimates reported in
Section~\ref{sec:h2_results} treat individual spatial tokens as
independent observations.
Because the $1{,}568$ tokens per video share the same semantic context,
they are not statistically independent; the effective sample size is
closer to the number of videos (approximately $9$--$10$ per category)
than to the number of tokens.
The reported $p$-values should therefore be interpreted as upper bounds
on statistical significance under token-level analysis rather than
definitive probability estimates.
A video-level analysis---treating each video's dominant symbol
assignment as the unit of observation---is deferred to future work due
to the small per-category sample size in the current experimental
setup.

As a supplementary analysis, we note that a label-shuffling permutation
test---randomly reassigning condition labels while preserving token
counts---would provide a stronger inferential baseline by sampling from
the actual joint distribution of the data.
Implementation of permutation testing is deferred to future work.

\section{Discussion}
\label{sec:discussion}

\subsection{Passive Probing as a Methodological Contribution}
We note that VQ is not a structurally neutral operation: codebook size, distance metric, and EMA update dynamics each constitute inductive biases that may amplify certain geometric directions in the latent space while suppressing others. The key inferential claim is not that the probe is bias-free, but that it introduces no task-specific supervision or semantic label that could produce spurious structure independently of the encoder.
The central methodological claim of this work is that the attribution
problem---the inability to cleanly separate the contributions of a
latent model from those of an attached interpretation component---can
be resolved by design rather than by post-hoc analysis.
Existing interpretability approaches for JEPA-style models attach
learned components to the encoder output: linear probes are trained to
decode target variables~\cite{alain2017probing}, language model heads
are fine-tuned to produce natural language
descriptions~\cite{assran2025vjepa2}, and pixel decoders are trained
to reconstruct visual inputs from latent vectors.
In each case, the interpretation component has its own learned
parameters, and when the composite system performs well it is
impossible to determine how much of the performance originates in the
encoder's representations versus the attached component's capacity.

The passive quantization design adopted here resolves this confound
structurally.
Because the V-JEPA~2 encoder is completely frozen
($\nabla_\phi \mathcal{L} = 0$) and the AIM quantizer operates
without any predefined semantic vocabulary, any systematic
relationship between the resulting symbol distributions and physical
conditions must originate in the encoder's pre-trained representations.
The quantizer cannot manufacture structure that the encoder does not
already possess, and the encoder cannot adapt to make the quantizer's
task easier.
This design contrasts with Discrete JEPA~\cite{baek2025discrete} and
NLoTM~\cite{wu2024nlotm}, which demonstrate that symbolic
representations \emph{can be learned} through joint training; our
results demonstrate that such structure \emph{can also be revealed}
without modifying the underlying model.

This distinction matters beyond the present work.
As JEPA-style world models are deployed in settings where their
internal representations inform downstream decisions---robot
manipulation planning, action anticipation, safety-critical
perception---the ability to audit those representations without
disturbing them becomes practically important.
Passive probing provides a methodology for doing so.

\subsection{The Compact Latent Space: Feature, Not Failure}

The most striking empirical observation in Stage~1 is that all five
action categories predominantly map to the same dominant codebook
entry ($s = 5$), yet the symbol \emph{distributions} over the eight
codebook entries are statistically distinguishable across all three
intervention pairs.
A naive reading of this result---that the codebook has collapsed
because a single symbol dominates---misses the structure that the
data actually reveal.

The correct interpretation is that V-JEPA~2's latent space is highly
compact: diverse action categories share a large common
representational core, and semantic differences are encoded as graded
distributional variations within that shared space rather than as
categorical boundaries between separate regions.
This compactness is not a failure of representational capacity.
It is the expected signature of a model trained to predict masked
spatiotemporal content rather than to maximize inter-class
separability.
A world model that has internalized the shared physically grounded variables
underlying diverse actions---gravity, inertia, human body kinematics,
spatial continuity---should represent those actions in an overlapping
region of latent space, not in disjoint clusters.

The differential sensitivity of V-JEPA~2's representations across the
three intervention variables provides additional support for this
interpretation.
The motion\_speed intervention (marching vs.\ archery; JSD $= 0.342$)
produces a substantially stronger signal than either grasp\_angle
(archery vs.\ bowling; JSD $= 0.190$) or object\_geometry
(flying\_kite vs.\ high\_jump; JSD $= 0.190$).
This ordering is consistent with V-JEPA~2's pretraining objective:
predicting masked future frames requires modeling the temporal dynamics
of motion, so representations are more sensitive to differences in
temporal structure (periodic gait vs.\ single aperiodic release) than
to differences in static object morphology or grip configuration.
The encoder has, in a sense, organized its latent space according to
what is most predictively informative---and AIM's symbol distributions
reflect that organization.

\subsection{Limitations and Boundary Conditions}

\paragraph{Discretization as a confound.}
The observed symbolic structure may partially reflect the VQ mechanism's tendency to impose cluster boundaries on smooth continuous distributions rather than latent structure pre-existing in the encoder's representation.
A continuous-space analysis---for example, applying kernel density estimation or Gaussian mixture modeling directly to the raw projected latent vectors---would provide evidence that the geometric structure exists prior to quantization.
Such analysis is deferred to Stage~2.

Three limitations of the Stage~1 results should be stated clearly.

\paragraph{Category-proxy confounding.}
The category-contrast experiments use action-category pairs as proxies for
physical-variable differences.
The symbol-distribution differences we observe therefore reflect the
overall latent-space distance between the selected categories, not the
isolated effect of the target physical variable.
For the grasp\_angle intervention, the latent distance between archery
and bowling simultaneously encodes differences in grip morphology,
scene environment, object material, and background color; the
experimental design provides no means of disentangling these
contributions.
The correct characterization of our results is: \emph{AIM symbols
have statistically significant discriminative power for action-category
pairs whose primary difference lies along a specified physical
dimension}---not that AIM symbols directly encode the isolated target
variable.
Establishing the latter claim requires video data in which the target
variable is manipulated while all other visual factors are held
constant, which is deferred to Stage~3 or future work.

\paragraph{Codebook resolution.}
With $K = 8$, the codebook is sufficient to detect the distributional
shifts reported in Stage~1, but too coarse to resolve fine-grained
intra-category variation.
The dominant-symbol collision is partly a consequence of this
resolution limit: a larger codebook would subdivide the cluster near
entry~\#5 into multiple entries, potentially revealing sub-structure
that is currently invisible.
Whether increasing $K$ produces qualitatively richer symbolic
distinctions or merely finer quantization of the same compact
structure is an empirical question addressed in Stage~2.

\paragraph{Statistical structure versus causal understanding.}
The category-contrast experiments establish that V-JEPA~2's latent
representations contain statistically structured manifolds that
correlate with physical conditions.
They do not establish that the encoder \emph{understands} those
conditions in any causal or mechanistic sense.
The observed symbol distributions are consistent with both a model
that has learned the causal structure of physical motion and a model
that has learned surface-level statistical regularities that happen to
correlate with physical variables.
Distinguishing between these interpretations requires interventional
experiments with synthetic or precisely controlled data, which are
beyond the scope of Stage~1.
The results suggest that latent representations contain extractable
structure, but not necessarily causal semantics.

\subsection{Implications for Latent Model Interpretability}
By `statistically testable interface' we mean a representation layer whose outputs can be subjected to standard hypothesis tests---chi-squared tests of distributional homogeneity, mutual information estimation, Jensen--Shannon divergence---without requiring human annotation or task-specific evaluation criteria.

The findings of Stage~1 have implications that extend beyond the
specific combination of V-JEPA~2 and AIM.

First, the passive probing methodology is not specific to video
encoders.
Any model that produces a continuous latent representation---a
reinforcement learning world model, a predictive coding
architecture~\cite{lecun2022path}, a multi-agent communication
system~\cite{liu2025aim}---is in principle amenable to the same
approach.
The prerequisite is only that the latent space be geometrically
structured enough for a vector quantizer to learn non-degenerate
codebook entries, a condition that Stage~1 has shown to be satisfied
for V-JEPA~2.

Second, the observation that V-JEPA~2's representations are compact
rather than categorically separated suggests that current latent
models may be better understood as systems that learn
\emph{shared physical abstractions} rather than categorical
classifiers operating in a high-dimensional space.
This framing is more consistent with the theoretical motivation for
JEPA~\cite{lecun2022path} than the standard evaluation practice of
measuring linear separability of class labels.
AIM's symbol distributions, which capture graded distributional
differences within a shared representational space, may be a more
appropriate tool for characterizing such models than probes designed
for categorical discrimination.

Third, the statistical auditability of discrete symbol sequences has
implications for AI safety applications.
Prior work has demonstrated that vector quantization can convert
unobservable latent communication between agents into auditable
statistical objects, enabling detection of covert
coordination~\cite{liu2026drcb}.
The present results suggest that the same principle applies to the
internal representations of world models: by routing latent vectors
through a discrete bottleneck, it becomes possible to monitor whether
a model's internal state is changing in ways that correlate with
external conditions, without requiring access to the continuous latent
vectors themselves.

\section{Future Work}
\label{sec:future}

The present work establishes Stage~1 of a four-stage research program
integrating AIM's discrete semantic layer with V-JEPA~2's latent
representations.
This section describes the planned trajectory of the remaining stages,
the extension of the framework to other latent model classes, and the
experimental designs required to address the causal limitations
identified in Section~\ref{sec:discussion}.

\subsection{Four-Stage Integration Roadmap}

The four-stage design is motivated by the progressive unfreezing
principle: the encoder starts completely frozen and is released only
after each preceding stage has established that the current
configuration is architecturally sound.
This sequencing ensures that any observed symbolic structure at each
stage is attributable to the representations that have been validated
in previous stages, rather than to co-adaptation between the encoder
and the quantizer.

\paragraph{Stage~1 (this work): Perception Gap Diagnosis.}
Goal: verify that AIM's symbolization mechanism can read structured
information from V-JEPA~2's frozen latent space without modifying the
encoder.
Status: complete.
All four pass criteria are satisfied
(H1 symbol stability $\bar{\rho} = 1.000$;
H2 $\chi^2\ p < 10^{-4}$ across all three category-contrast experiments;
absolute MI $0.036$--$0.117$\,bits, normalized MI $1.2$--$3.9\%$;
codebook active ratio $= 62.5\%$).
The encoder weights remain completely frozen throughout.

\paragraph{Stage~2: Codebook Stabilization and Scaling.}
Goal: establish a richer, more stable symbolic vocabulary on the
frozen encoder.
Primary changes: increase codebook size $K$ from 8 to 32 or 64,
introduce residual vector quantization to resolve sub-structure within
the dominant latent cluster identified in Stage~1, and train on a
larger and more diverse video dataset to improve codebook coverage.
Entry condition: Stage~1 pass criteria satisfied (met).
Pass criteria for Stage~3 entry: codebook active ratio $> 60\%$
across all $K$ entries; intervention JSD $> 0.3$ for all three
physical variables; intra-class symbol consistency $> 80\%$ (a
stronger stability criterion than the pipeline determinism check used
in Stage~1).
A key diagnostic question for Stage~2 is whether increasing $K$
produces qualitatively richer symbol differentiation---supporting the
codebook-resolution interpretation of dominant symbol collision---or
merely finer quantization of the same compact representational
core---supporting the latent-compactness interpretation.
The differential JSD sensitivity already observed in Stage~1
(motion\_speed JSD $= 0.342$, $1.8\times$ higher than the other two
interventions) provides a falsifiable prediction: if latent compactness
is the primary explanation, the JSD ordering across physical dimensions
should be preserved at larger~$K$.

\paragraph{Stage~3: Symmetric Quantization and Joint Training.}
Goal: allow the encoder to adapt its representations toward the
symbolic vocabulary established in Stage~2.
The encoder is unfrozen for the first time.
A two-phase warm-up strategy is required: first, allow the encoder to
stabilize in continuous latent space with the quantizer loss held at
zero; then gradually introduce the VQ commitment loss using a
$\beta$-schedule, so that the encoder learns to produce representations
that are more amenable to discrete symbolization without losing the
physically grounded variables acquired during pretraining.
A Teacher--Student asymmetry must be managed carefully: because
V-JEPA~2's Teacher encoder is an EMA of the Student, introducing VQ
quantization on both sides simultaneously creates a moving-target
instability.
The recommended approach is to quantize only the Student path in the
first sub-stage and extend to the Teacher path only after the Student
codebook has converged.
Entry condition: Stage~2 pass criteria satisfied.

\paragraph{Stage~4: Action-Conditioned Symbolic World Model and
Causal Intervention.}
Goal: extend the symbolic interface to action-conditioned prediction
and validate causal structure through controlled physical
interventions.
Building on V-JEPA~2-AC~\cite{assran2025vjepa2}, an action-conditioned
predictor will be trained on top of the jointly-trained encoder from
Stage~3.
The resulting system should produce symbol sequences that change
predictably in response to action inputs, enabling a form of
symbolic planning: given a goal state expressed as a target symbol
distribution, the system can evaluate candidate action sequences by
predicting their symbolic consequences.
Causal validation will use synthetic or robot-manipulation footage
with precise physical control, isolating individual physical variables
while holding all other visual factors constant---the experimental
design required to move from the statistical correlations established
in Stage~1 to genuine causal claims.

\subsection{AIM as a General Semantic Interface for Latent Models}

The three-layer framework introduced in
Section~\ref{sec:framework}---latent model, discrete semantic layer,
language interface---is not specific to V-JEPA~2.
The same architecture applies to any model that produces a continuous
latent representation of sufficient geometric structure.

In the context of reinforcement learning world
models~\cite{lecun2022path}, AIM's discrete layer would serve as a
symbolic state representation: rather than operating on continuous
latent state vectors, a planning algorithm could reason over symbol
sequences, which are finite, auditable, and composable.
This has potential implications for explainability in safety-critical
applications, where human operators need to understand why a model
took a particular action.

In multi-agent systems, AIM's original application
domain~\cite{liu2025aim}, the discrete semantic layer provides a
shared symbolic vocabulary through which agents can communicate about
their internal states without requiring a pre-defined communication
protocol.
The present work demonstrates that this vocabulary can be grounded in
a pre-trained world model's representations, providing a path toward
agents that communicate about physically meaningful concepts rather
than arbitrary learned codes.
The DRCB framework~\cite{liu2026drcb} has further shown that
discrete symbol sequences can be monitored for anomalous distributional
drift, enabling detection of covert coordination.
Combining these capabilities---world-model grounding, inter-agent
communication, and safety monitoring---represents a medium-term
research goal for the AIM program.

\subsection{Language Interface Layer}

Stage~1 through Stage~3 develop the first two layers of the
three-layer framework: the latent model and the discrete semantic
layer.
Stage~4 and beyond will address the third layer: a language model
that maps discrete symbol sequences into natural language descriptions
of the model's internal state.

Two design principles will guide this development.
First, the language model should operate on symbol sequences rather
than on continuous latent vectors, preserving the attribution property
established in Stage~1: any natural language description produced by
the system is grounded in the symbolic record, not in the language
model's own priors about what the encoder ``should'' have seen.
Second, the symbol-to-language mapping should be evaluated not by
fluency or downstream task accuracy, but by causal fidelity: does
the natural language description change in the expected direction when
a physical variable is manipulated?
This evaluation criterion connects the language interface layer
directly to the intervention-based methodology developed in Stage~1.

\subsection{Toward Causal Validation}

The Stage~1 results establish statistical associations between AIM
symbol distributions and physical conditions.
Establishing causal claims requires a stronger experimental design
in which the target variable is manipulated while all other visual
factors are held constant.

Three directions are planned.
First, synthetic video datasets generated under controlled physical
simulation will allow precise manipulation of individual variables
(object mass, surface friction, gravitational acceleration) while
holding scene appearance constant.
Second, robot-manipulation footage from the Franka arm datasets used
in V-JEPA~2-AC~\cite{assran2025vjepa2} provides a real-world setting
with a high degree of physical control.
Third, counterfactual probing---presenting the encoder with video
pairs that differ in a single physical event (e.g., a collision that
does or does not occur)---will test whether the symbol distributions
respond to causal structure rather than surface-level statistical
regularities.

Together, these directions constitute the experimental foundation for
Stage~4, and for the broader claim that AIM's discrete symbol system
can serve as a causally grounded interpretability interface for
video world models.

\newpage

\bibliographystyle{unsrt}
\bibliography{references}

\end{document}